\documentclass[]{style/iromlab}

\DeclareDocumentEnvironment{example}{}{\noindent\textbf{Running example:}\itshape}{}

\usepackage{ifthen}
\newboolean{include-notes}
\newboolean{include-new}
\newboolean{include-remove}
\setboolean{include-notes}{true}
\setboolean{include-new}{false}
\setboolean{include-remove}{false}

\usepackage[dvipsnames]{xcolor}
\usepackage[normalem]{ulem}
\providecommand{\justin}[1]{\ifthenelse{\boolean{include-notes}}{\textcolor{orange}{\textbf{Jaime:} #1}}{}}

\providecommand{\princeton}[1]{\ifthenelse{\boolean{include-notes}}{\textcolor{orange}{#1}}{}}

\providecommand{\p}[1]{\smallskip \noindent \textbf{{#1}.}}

\providecommand{\mcal}[1]{\mathcal{#1}}
\providecommand{\mbb}[1]{\mathbb{#1}}

\providecommand{\mbf}[1]{\ensuremath{\mathbf{#1}}}
\providecommand{\Expect}{\ensuremath{\mathbb{E}}}

\newcommand*{\de}{\ensuremath{\mathrm{d}}\xspace}

\usepackage{multicol}
\usepackage{amsmath, amsfonts, amssymb, amsthm}
\usepackage{bm}
\usepackage[inline]{enumitem}
\usepackage{mathtools}
\usepackage{graphicx}
\usepackage{longtable,tabularx}
\usepackage{placeins} 
\usepackage{float}
\usepackage{multirow}
\usepackage{bbm}
\usepackage{threeparttable}
\usepackage{balance}
\usepackage[ruled,algo2e]{algorithm2e}
\usepackage{algpseudocode}
\usepackage{adjustbox}
\usepackage{booktabs}
\usepackage{bm}
\usepackage{etoolbox}
\usepackage{microtype}
\usepackage[title]{appendix}
\usepackage{units}
\usepackage{cleveref}
\usepackage{xspace}
\usepackage{tcolorbox}
\usepackage{wrapfig}
\usepackage{caption}
\usepackage{thm-restate}
\usepackage{lineno}
\usepackage{makecell}
\usepackage{lmodern}

\usepackage{soul}
\setuldepth{foobar}

\crefname{section}{Sec.}{Secs.}
\crefname{figure}{Fig.}{Figs.}
\crefname{table}{Tab.}{Tabs.}
\crefname{equation}{Eq.}{Eqs.}

\Crefname{section}{Sec.}{Secs.}
\Crefname{figure}{Fig.}{Figs.}
\Crefname{table}{Tab.}{Tabs.}
\Crefname{equation}{Eq.}{Eqs.}

\newbool{extended}
\setbool{extended}{false}

\makeatletter
\providecommand{\longdash}[1][2em]{%
  \makebox[#1]{$\m@th\smash-\mkern-7mu\cleaders\hbox{$\mkern-2mu\smash-\mkern-2mu$}\hfill\mkern-7mu\smash-$}}
\makeatother
\providecommand{\omitskip}{\kern-\arraycolsep}

\providecommand{\hugeLARGE}{\fontsize{20pt}{22pt}\selectfont}

\definecolor{tbox_color}{HTML}{78B6E8}

\tcbset{
  tbox_style/.style={
    colback=tbox_color!10,
    colframe=tbox_color!60,
    coltitle=black,
    boxrule=1.0pt,
    arc=2pt,
    outer arc=2pt,
    left=4pt,
    right=4pt,
    top=4pt,
    bottom=4pt,
    fonttitle=\bfseries,
  }
}

\makeatletter
\newcommand{\listofquestions}{\section*{List of Questions}\@starttoc{loq}}
\newcounter{question}

\newcommand{\qitem}[3][]{%
  \refstepcounter{question}%
  \ifx\relax#1\relax
    \label{q:\thequestion}%
  \else
    \label{#1}%
  \fi
  \item[Q\thequestion:] #2%
  \addcontentsline{loq}{section}{\textbf{Q\thequestion:} #2}%
  \par\noindent\textbf{Answer:} #3%
}
\makeatother
\makeatother

\providecommand{\algname}{CLAP\xspace}
\providecommand{\bfalgname}{\mbox{\textbf{\algname}}\xspace}
\providecommand{\alglam}{\texttt{CLAP-LAM}\xspace}
\providecommand{\algee}{\texttt{CLAP-EE}\xspace}
\providecommand{\alglang}{\texttt{CLAP-LANG}\xspace}
\providecommand{\algcurr}{\texttt{CLAP-CURR}\xspace}

\usepackage{enumerate}
\usepackage{cleveref}
\usepackage{titletoc} %


\author[*]{Kechen Liu}
\author[*]{Ola Shorinwa}

\affiliation[]{Princeton University}
\contribution[*]{Equal contribution.}

\begin{document}

\title{
{\hugeLARGE
\algname:
Cross-Embodiment Video World Models are Zero-Shot Physical Simulators}
}

\abstract{
    State-of-the-art action-conditioned video models are typically restricted to a single robot embodiment, preventing them from leveraging the vast corpus of heterogeneous video data that contains rich signals for learning generalizable physics. To bridge this gap, we introduce \algname, a framework for cross-embodiment action-conditioned video generation capable of being trained on diverse, internet-scale videos across human and robotic agents. \algname is grounded in the insight that universal physical laws govern spatiotemporal dynamics regardless of the actor. However, cross-embodiment learning is non-trivial because action representations vary sharply across robot platforms and are typically absent in human videos. \algname addresses  this fundamental challenge through three core contributions. 
First, \algname reconciles disparate action spaces across human and robot morphologies using end-effector poses, natural language instructions, and learned latent action representations. 
Second, to resolve the individual limitations of each action representation, \algname introduces a curriculum-based cross-embodiment learning recipe that first learns foundational physical priors across unlabeled video data using latent actions and subsequently grounds them in end-effector action spaces for zero-shot deployment to real-world tasks.
Crucially, \algname approaches or surpasses state-of-the-art single-embodiment video models in challenging environments like DROID, with performance advantages that compound via few-shot adaptation to target embodiments. Through inference-time cross-policy planning and reinforcement-learning-based policy finetuning in video world models, we demonstrate \algname's zero-shot generalization to real-world tasks, improving the performance of state-of-the-art robot policies, such as $\pi_{0.5}$ and MolmoAct-2. Ultimately, \algname delivers the most comprehensive suite of action-conditioned video world models to date --- spanning diverse action-conditioning spaces (end-effector, language, and latent) and robot morphologies (including cross-embodiment, DROID, Bridge, bimanual YAM robots, and G1~humanoids).
Third, \algname establishes a novel paradigm for training high-fidelity single-embodiment video world models via sample-efficient, few-shot adaptation of cross-embodiment models to target robot platforms.
We open-source all code and models at \href{https://github.com/omni-CLAP/clap}{{\color{irom_red} https://github.com/omni-CLAP/clap}}.
}


\website{
https://omni-clap.github.io  %
}
{
omni-clap.github.io   %
}

\code{
https://github.com/omni-CLAP/clap %
}
{
github.com/omni-CLAP/clap  %
}

\maketitle

\section{Introduction}
\label{sec:intro}
The defining breakthroughs in Large Language Models (LLMs) were precipitated by web-scale training on diverse text datasets, shattering performance barriers by learning a unified representation of human language~\citep{touvron2023llamaopenefficientfoundation, chowdhery2022palmscalinglanguagemodeling, kaplan2020scalinglawsneurallanguage, brown2020languagemodelsfewshotlearners}. Today, as action-conditioned video world models continue to struggle with physical inconsistency, the robotics and vision communities face a critical question: can generative video models learn generalizable physical priors through cross-embodiment data scaling, or does the heterogeneous nature of different robot morphologies demand agent-specific architectures? In this work, we investigate how video world models can bridge the gap between diverse embodiments and analyze the structural bottlenecks that must be overcome.

Prior action-conditioned video models~\citep{guo2026ctrlworldcontrollablegenerativeworld, quevedo2025worldgym} focus exclusively on single robot embodiments, sidestepping the inherent challenges associated with diverse robot morphologies, which forecloses the internet-scale video data behind recent foundation-model advances. To address these limitations, we introduce \bfalgname, a training framework for cross-embodiment action-conditioned video models designed to learn fundamental physical laws across different agents. \algname's core insight lies in the fact that the same universal laws of physics govern all dynamical interactions irrespective of the agent's embodiment. For example, a bottle tips, a towel folds, and a drawer slides
according to the same physics whether the actor is a Franka arm, a WidowX gripper, or a human
hand, underscoring that the appropriate response to heterogeneity is not to restrict training to one embodiment. 

Translating this physical principle into a scalable framework, however, requires overcoming severe cross-embodiment challenges. Web-scale datasets feature actors with vastly different physical morphologies and action spaces, and crucial subsets (e.g., human videos) are entirely devoid of action labels. 
\algname bridges this gap via three fundamental contributions.
First, \algname harmonizes the disparate action spaces of diverse embodiments using end-effector poses, natural language instructions, and learned latent action representations. However, these action representations have native limitations. For example, while latent actions enable learning from unlabeled videos, they do not facilitate zero-shot real-world deployment. Conversely, while end-effector actions address this drawback, they preclude training on unlabeled video data. 
Second, to address these limitations, \algname introduces a curriculum-based cross-embodiment learning framework that first learns foundational physical priors from unlabeled video data using latent actions and subsequently refines them in end-effector action spaces for direct, zero-shot deployment to real-world tasks.
Third, \algname establishes a novel paradigm for training single-embodiment video models through sample-efficient adaptation of cross-embodiment models to target embodiments. This adaptation framework facilitates the seamless transfer of physical priors across highly disparate robot morphologies.

Critically, \algname approaches or surpasses state-of-the-art single-embodiment baselines in challenging environments like DROID~\cite{khazatsky2025droidlargescaleinthewildrobot}. Notably, this performance advantage compounds through sample-efficient adaptation to the target embodiment. Through inference-time cross-policy planning and reinforcement-learning-based policy finetuning in video world models, we demonstrate \algname's zero-shot generalization to real-world manipulation tasks, achieving higher success rates compared to the base performance of state-of-the-art policies, e.g., $\pi_{0.5}$~\cite{intelligence2025pi} and MolmoAct-2~\cite{fang2026molmoact2actionreasoningmodels}.

\begin{figure*}[t]
    \centering
    \includegraphics[width=\linewidth]{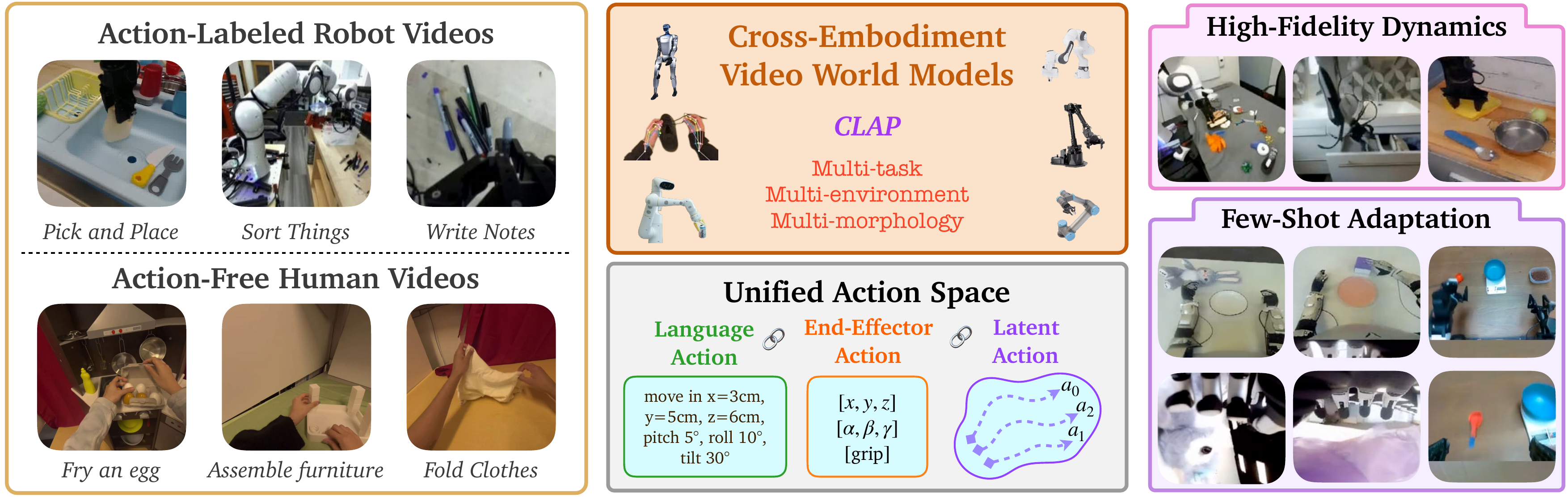}
    \caption{We introduce \textbf{CLAP}, a cross-embodiment learning framework that trains action-conditioned video models on diverse human and robot video data to learn fundamental physical priors for fine-grained dynamics prediction.}
    \label{fig:teaser}
    \vspace{-2ex}
\end{figure*}

Ultimately, \algname delivers the most comprehensive suite of action-conditioned video world models to date (\Cref{fig:teaser}), which spans diverse action-conditioned spaces (end-effector, language, and latent) and robot morphologies (e.g., cross-embodiment, DROID, Bridge, bimanual YAM robots and G1~humanoids). We open-source all video world models with accompanying code. (See Appendix~\ref{sec:app_prelims} for preliminaries and Appendix~\ref{sec:app_nuanced_summary} for a nuanced summary of the core contributions of \algname.)

\section{Related Work}
\label{sec:related_work}

\p{Video Generation in Robotics}
Recent advances in video generation models~\citep{agarwal2025cosmos,wan2025,blattmann2023stable,brooks2024sora} have enabled high-fidelity synthesis of physically consistent video content, motivating their growing adoption as embodied world models in robotics. One line of work leverages video models as data generators, synthesizing robot trajectories paired with pseudo-action labels for downstream policy learning~\citep{jang2025dreamgen,bharadhwaj2024gen2act}. Other works directly employ video models as policy backbones, decoding actions through tracking, inverse dynamics, or unified video-action prediction~\citep{du2024video,hu2024video,liao2025genie,li2025unified,zhu2025unified} (see~\citep{mei2026video} for a review of these video generation models). Most relevant to our work, a growing body of research employs action-conditioned video models as learned simulators for policy evaluation and improvement, rolling out policies in imagination space rather than on real hardware~\citep{quevedo2025worldgym,guo2026ctrlworldcontrollablegenerativeworld,jiang2024irasim,tseng2025scalable,gemini2025veo,jain2026weaverbetterfasterlonger,yin2026playworldlearningrobotworld}. However, all these methods are primarily restricted to a single robot embodiment, hindering them from leveraging diverse cross-embodiment video data. Our work addresses these fundamental limitations.

\p{Cross-Embodiment Robot Learning and Latent Actions}
A central challenge in scaling generalist robot policies is the heterogeneity of robot embodiments, which exhibit varying degrees of freedom, kinematic structures, and observation configurations. To bridge this heterogeneity, prior work has explored several strategies for unifying state and action spaces across embodiments. The most direct approaches pad states and actions to a maximum dimensionality or adopt a shared end-effector pose or language representation~\citep{kim2024openvla, black2024pi0, zheng2025xvla, zha2026lap}, while others learn embodiment-specific action heads or projectors atop a shared backbone~\citep{octo2024,wang2024hpt,nvidia2025gr00t,doshi2024crossformer}.
Other approaches~\citep{bu2025univla, zheng2025uniact, liang2025clam, chen2025villax} leverage latent actions as a paradigm for unifying heterogeneous action spaces.
Genie~\citep{bruce2024genie} learns latent actions via a VQ-VAE objective for interactive video generation, while LAPA~\citep{ye2024lapa} applies them to VLA policies.
However, current latent action models (LAMs) typically require downstream alignment  to embodiment-specific commands. To address this challenge, we propose end-effector and language actions as unified action representations that directly leverage ground-truth actions for fine controllability when dense action-labeled video data is available. 
To combine their complementary strengths, \algname uses a curriculum-based video pretraining recipe that integrates latent actions with end-effector actions for direct, zero-shot deployment. 

A few concurrent studies~\cite{wu2026oscaromniembodimentactionconditionedworld, alzayer2026maskedvisualactionsunified, li2026hydra0actionflowgeneralist} have emerged during the preparation of this work that leverage privileged information --- such as robot URDFs, simulators, and camera intrinsics and extrinsics --- to largely bypass fundamental challenges in video modeling. Using these inputs, these approaches convert raw observations into explicit geometry signals, such as per-frame robot skeletons~\cite{wu2026oscaromniembodimentactionconditionedworld}, segmentation masks~\cite{alzayer2026maskedvisualactionsunified}, or pixel flow~\cite{li2026hydra0actionflowgeneralist}. These engineered shortcuts explicitly localize the robot or separate dynamic regions from passive areas in each future frame, effectively outsourcing some of the hardest parts of future prediction.
This reliance introduces critical limitations. In practical deployments, such privileged information is notoriously difficult to obtain and maintain (e.g., accurate camera calibration), and simulator dependence runs counter to the broader vision of video models serving as general-purpose replacements for traditional physics engines.
Furthermore, some of these geometric shortcuts impose strict visibility assumptions, requiring robot joints and end-effectors to remain continuously in-frame, a constraint frequently violated in real-world settings.
\algname avoids these limitations entirely, confronting the video modeling challenge directly by requiring only the initial camera observation and the desired or delta 3D location as per-frame actions.

\section{CLAP: Training Cross-Embodiment Video World Models}
\label{sec:method}
\begin{wrapfigure}[12]{r}{0.5\columnwidth}
    \vspace{-3ex}
    \centering
    \includegraphics[width=\linewidth]{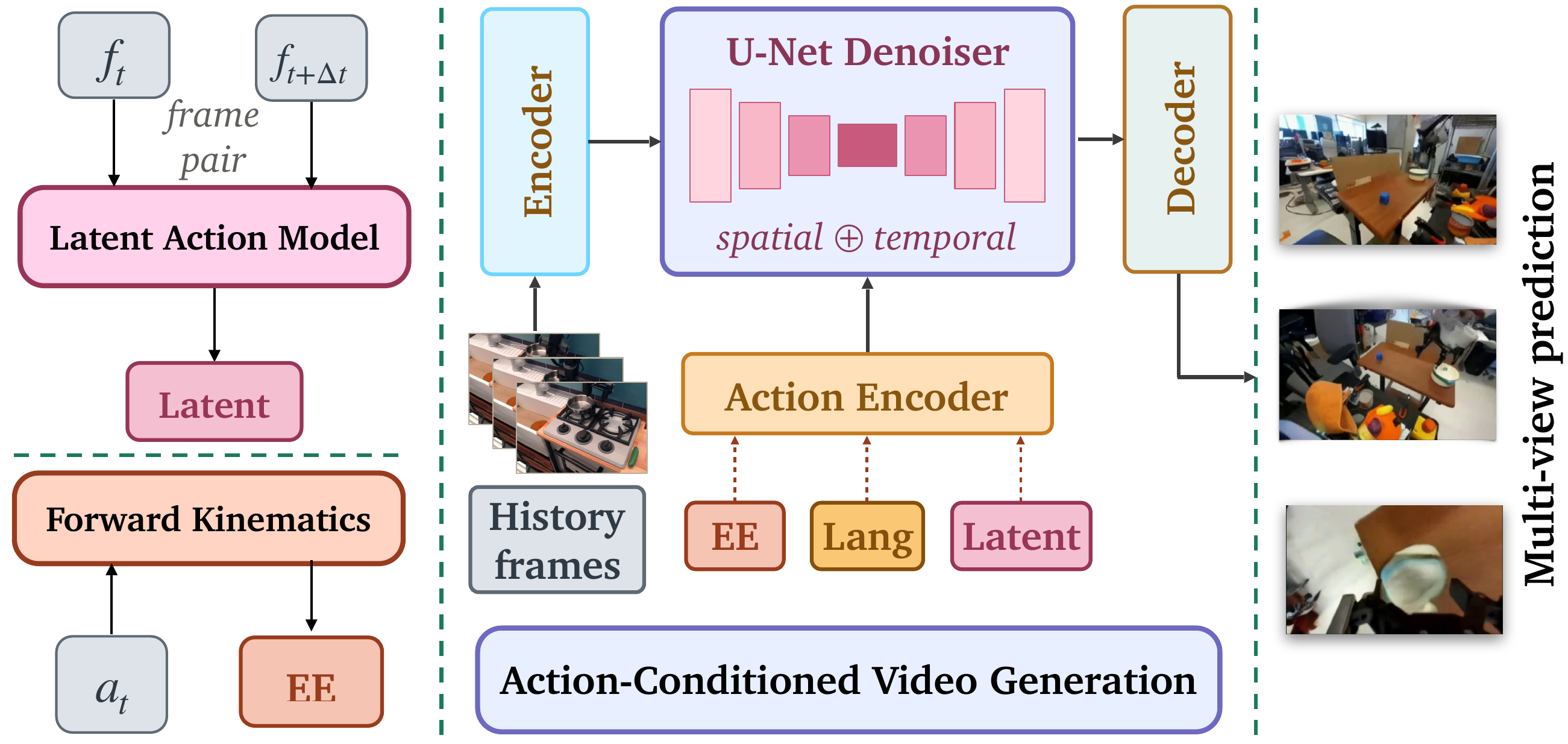}
    \caption{\textbf{\algname} \algname reconciles disparate action spaces across human and robot morphologies to train high-fidelity cross-embodiment video models.}
    \label{fig:pipeline}
\end{wrapfigure}
Cross-embodiment learning from video data presents a unique challenge:
robot and human data is inherently heterogeneous with different observation and action spaces, and internet-scale video data largely lacks action labels. 
To address these challenges, \algname introduces a novel recipe for training cross-embodiment video world models, reconciling heterogeneous action spaces using end-effector poses, natural language, and latent action representations (see~\Cref{sec:method_action_harmonization,sec:method_pre_training}). 
\algname facilitates data-efficient, few-shot adaptation to target embodiments (see~\Cref{sec:method_post_training}). \Cref{fig:pipeline} summarizes this pipeline.

\subsection{Harmonizing Heterogeneous Action Spaces}
\label{sec:method_action_harmonization}
Varying robotic degrees of freedom (DoF) dictate highly divergent action spaces; for instance, the ubiquitous WidowX and Franka platforms operate with six and seven DoFs, respectively. Because action-conditioned video models require fixed-dimensional conditioning inputs, generalizing across these diverse morphologies demands a unified action space.
To resolve this gap, \algname establishes unified action representations in end-effector, language, and latent action-spaces and applies critical transformations to bridge the disparate action spaces of diverse robot embodiments. While each action representation offers unique advantages, they also introduce distinct limitations. Below, we analyze these trade-offs and introduce a curriculum-based action harmonization method that combines the strengths of different action representations to overcome their individual drawbacks.
We provide additional details in Appendix~\ref{app:action_representations}.

\p{End-effector Actions} 
While joint-space control is widely adopted by prior work in robot control, it inherently restricts a video model’s amenability to architectures sharing identical degrees of freedom. To facilitate cross-embodiment learning, we introduce \algee, which adopts end-effector (EE) actions as unified, task-relevant representations that abstract away low-level morphological differences. Concretely, we map joint positions to a 7-DoF operational space via the robot's forward kinematics. In line with prior work~\cite{kim2024openvla}, our canonical end-effector action space comprises three translational dimensions and three Euler angles for orientation, alongside a continuous gripper state. This representation is readily compatible with existing large-scale data repositories, such as the Open X-Embodiment (OXE) dataset~\citep{o2024open}, which natively provide frame-level joint position and kinematic metadata.
Since the volume of a robot's operational space varies widely with its morphology, end-effector actions demand video modeling across larger action spaces, which typically poses an intractable challenge, particularly in absolute-action spaces. To address this limitation, prior work~\cite{gao2026dreamdojo} trains single-embodiment video models using relative actions. However, we show that relative-action spaces degrade the prediction fidelity of cross-embodiment video models conditioned on end-effector actions (see~\Cref{sec:exp_rel_vs_abs_action_spaces}). To circumvent this degradation, \algee utilizes absolute end-effector actions and normalizes the disparately-sized action spaces of each robot platform into finite bounds ${[-1, 1]}$, effectively simplifying the learning problem within a shared, uniformly-bounded action space across all robots. Crucially, each robot inherits a unique normalization factor, accounting for its morphological differences. Although these action harmonization architecture overcomes the aforementioned challenges, a central limitation persists: end-effector action conditioning requires access to labeled video data, impeding data scaling to the massive corpus of unlabeled internet videos.

\p{Language Actions} 
Language actions express control primitives as natural text, offering a versatile abstraction that enables coarse-to-fine controllability with minimal architectural adaptation. Furthermore, recent work~\citep{hancock2025actions} suggests that text-based action abstractions are more effective in preserving pre-trained spatiotemporal priors within foundational models. Therefore, we derive \alglang, a recipe for training cross-embodiment models conditioned on language actions.
To maintain controllability within a textual interface, we project numerical end-effector actions into the language space by mapping continuous coordinate actions to text. Preliminary experiments revealed that the high-precision format of ground-truth robot end-effector actions often leads to truncation by the CLIP tokenizer. Moreover, existing state-of-the-art text tokenizers are  poorly aligned to preserve the uniqueness of numerical values during tokenization, potentially compromising the information content of tokenized actions. \alglang uses concise templated motion primitives (e.g., ``$x$={}, $y$={}, ...'') to compensate for the limited context length of text tokenizers and the limited resolution of the discrete language-tokenization space. In addition to normalizing the action space, \alglang transforms absolute end-effector actions into relative-action spaces to further narrow the action space, improving the information density of templated language actions. We demonstrate that relative-action spaces improve future-prediction fidelity in cross-embodiment video models in~\Cref{sec:exp_rel_vs_abs_action_spaces}. However, compensating for the weaknesses of language tokenization still requires decreasing the precision of the conditioning actions, ultimately limiting the spatial resolution of language-action-conditioned video models compared to their end-effector-action-conditioned counterparts.

\p{Latent Actions}
Large volumes of web-scale video data do not contain action annotations, particularly human videos, impeding their amenability to end-effector and language actions. To harness the rich diversity of unlabeled video data in training cross-embodiment video models, we leverage latent actions as proxy action representations, which describe the underlying transformation between pairs of video frames using (low-dimensional) action tokens. Latent action models (LAMs)~\citep{ye2024lapa, bruce2024genie} utilize autoencoders (VAEs) to learn pseudo-actions from videos through self-supervision on a reconstruction loss objective. \algname's LAM takes in a pair of video frames ${(f^{t}, f^{t+\Delta t})}$ (${\Delta \in \mbb{Z}}$) and outputs $d$-dimensional continuous latent actions (we set ${d = 32}$). The encoder ${q_{\phi}}$ consists of a spatiotemporal transformer that extracts features from both input video frames, which is mapped to latent action ${a_{\text{ln}} \in \mbb{R}^{d}}$. To reconstruct $f^{t+\Delta t}$, the decoder ${p_{\psi}}$ takes in $f^{t}$ and $a_{\text{ln}}$, incentivizing the model to learn causal relationships. We train the LAM with the reconstruction loss:
\begin{equation}
    \label{eq:lam_objective}
    \mcal{L}_{\psi, \phi}^{\text{LAM}}(f^{t + \Delta t}) = \Expect_{q_{\phi}(a_{\text{ln}} \mid f^{t}, f^{t+\Delta t})} \log p_{\psi}(f^{t+\Delta t} \mid f^{t}, a_{\text{ln}}) - \alpha D_{\text{KL}}(q_{\phi}(a_{\text{ln}} \mid f^{t}, f^{t+\Delta t}) \Vert p_{l}(a_{\text{ln}})),
\end{equation}
with KL divergence-regularization, where ${\alpha \in \mbb{R}_{+}}$ denotes the relative weight between the two objective function terms, and ${p_{l}(a_{\text{ln}})}$ represents the prior distribution. 
During training, we dynamically vary $\Delta t$ as a form of data augmentation to improve model robustness. 
We train the video model \alglam using the latent actions extracted from raw (unlabeled) video data for effective cross-embodiment scaling, which can improve prediction accuracy (see~\Cref{sec:exp_cross_vs_single}). Despite this strength, \alglam requires downstream alignment to a target embodiment's action space, which can degrade its prediction fidelity, as observed in preliminary experiments.

\p{Curriculum-based Latent-to-EE Actions}
\algname's end-effector-action, language-action, and latent-action harmonization methods offer valuable unification advantages; however, they all face important challenges.
To resolve these limitations, we introduce \algcurr, a curriculum-based video pretraining method that first learns foundational physical priors from unlabeled video data using latent actions and subsequently refines these priors conditioned on end-effector actions for zero-shot deployment to real-world tasks.
Critically, \algcurr combines the unique strengths of latent actions with those of end-effector actions to overcome their individual limitations. This strategy achieves unrestricted cross-embodiment scaling without requiring downstream adaptation to target robot embodiments. Moreover, by eliminating the need for downstream adapters, \algcurr circumvents the degradation in prediction fidelity typically introduced by alignment layers. After training the video model with latent actions, \algname swaps the action head with another that takes in $7$-dimensional end-effector actions, while retaining the pretrained weights in the video model backbone. Thereafter, \algcurr jointly trains the new action head and the video model on action-labeled video data.

\subsection{Training the Cross-Embodiment Video World Model}
\label{sec:method_pre_training}

\p{Model Architecture}
\algname uses the video diffusion~\citep{karras2022elucidatingdesignspacediffusionbased} paradigm for action-conditioned video generation, which is described by the probability flow ordinary differential equation: ${\de\mbf{x} = -\dot{\sigma}(t) \sigma(t) \nabla_{\mbf{x}} \log p_{v} (\mbf{x}; \sigma(t)) \de t,}$ given the input video $\mbf{x}$ sampled from the probability distribution $p_{v}$ and noise scheduler $\sigma(t)$. We supervise the video model $\mcal{V}_{\theta}$ using the loss function:
\begin{equation}
    \label{eq:loss_fn}
        \Expect_{\sigma, \mbf{x}_1, \mbf{x}_0} \Bigl[w(\sigma) \Bigl\Vert \mcal{V}_{\theta}(c_{\mathrm{in}}(\sigma) \cdot \mbf{x}_{c}; c_{\mathrm{noise}}(\sigma))  %
         - \frac{1}{c_{\mathrm{out}}(\sigma)}(\mbf{x}_1 - c_{\mathrm{skip}}(\sigma) \cdot \mbf{x}_{c}) \Bigr\Vert_{2}^{2} \Bigr],
\end{equation}
with partially denoised video input ${\mbf{x}_{c} = \mbf{x}_1 + \sigma \mbf{x}_0}$, ground-truth video ${\mbf{x}_1 \sim p_{\mathrm{data}}}$, pure noise ${\mbf{x_0} \sim \mcal{N}(\mbf{0}, \mbf{I})}$, and scaling and conditioning factors ${(c_{\mathrm{in}}, c_{\mathrm{out}}, c_{\mathrm{skip})}}$ and $c_{\mathrm{noise}}$, where ${w(\sigma) = \lambda(\sigma) c_{\mathrm{out}}(\sigma)^{2}}$ and  ${\sigma \sim p_{d}}$. In practice, we train latent video diffusion models using continuous VAEs for video tokenization followed by a U-Net for salient spatiotemporal modeling.

\p{History Conditioning and Multi-view Video Prediction}
Standard action-conditioning inputs lack the temporal velocity information crucial for high-fidelity future prediction. To provide rich temporal context, we condition the video model on a history of past observations and robot poses across a horizon $H$, employing causal cross-attention to bridge historical context with future predictions. Furthermore, to enforce geometrically consistent training supervision and enhance downstream robustness, we incorporate multi-view conditioning to generate multi-view future frames. Structurally, we concatenate all views vertically in the latent space, passing the unified frame directly to the model without altering its core architecture.

\subsection{Beyond Zero-Shot: Data-Efficient Adaptation to Target Morphologies}
\label{sec:method_post_training}
Through broad-scale training on diverse robot video data, \algname's models learn generalizable priors that deliver high-accuracy dynamics prediction. Nonetheless, their zero-shot prediction accuracy can be improved by finetuning them on target embodiments, spanning pretrained embodiments which have been seen during training and novel embodiments.
Via finetuning, these models can refine their learned priors to better align them with target embodiments.

\p{Pretrained Embodiments}
\algname establishes a novel framework for training single-embodiment video world models via few-shot adaptation of cross-embodiment models. 
\algname's models provide foundational physical priors that can be readily adapted to a target embodiment via data-efficient finetuning on the robot's native end-effector action space. This finetuning process does not require any modifications to the architecture of the underlying cross-embodiment model, since the target embodiment's action space already matches that of the base model. Crucially, this paradigm for training high-fidelity single-embodiment video models yields superior world models compared to prevailing approaches that train them from scratch or from less-aligned video model backbones such as WAN~\cite{wan2025} and SVD~\cite{blattmann2023stable}.

\p{Novel Embodiments}
Beyond pretrained embodiments, \algname facilitates data-efficient adaptation of cross-embodiment models to novel robot morphologies that are markedly different from those seen during training, such as bimanual robots and humanoids, which typically feature higher-dimensional action spaces. Due to this action-space disparity, model finetuning requires substituting the model's action head with a new one compatible with the target embodiment.
Crucially, \algname retains all other components of the cross-embodiment video model to preserve rich spatiotemporal priors. During finetuning, \algname aligns these underlying priors with the target morphology, enabling high-fidelity action-conditioned future prediction.

\section{Experiments}
\label{sec:evaluation}
Through the lens of the following research questions, we investigate the impact of cross-embodiment scaling on action-conditioned video models and their efficacy in downstream robotics tasks:
\begin{itemize}[leftmargin=0pt, labelwidth=0pt, itemindent=0pt]
    \item[] \textbf{[Q1.]}~Do cross-embodiment video models surpass their single-embodiment counterparts?~(\Cref{sec:exp_cross_vs_single})
    \item[] \textbf{[Q2.]}~Do relative-action spaces outperform absolute-action spaces in cross-embodiment video modeling?~(\Cref{sec:exp_rel_vs_abs_action_spaces}) 
    \item[] \textbf{[Q3.]}~Do cross-embodiment video models achieve zero-shot real-world generalization?~(\Cref{sec:exp_real_world})
    \item[] \textbf{[Q4.]}~Beyond Zero-Shot: Does \algname facilitate data-efficient adaptation to target embodiments?~(\Cref{sec:exp_few_shot_adaptation}) 
\end{itemize}

\p{Evaluation Setup}
We train \algname's cross-embodiment models on the Open X-Embodiment~\cite{o2024open} and EgoDex~\cite{hoque2026egodex} datasets for 100K steps on eight H100 or H200 GPUs (varying by availability), taking approximately two to three days per training run. We use an effective batch size of $64$ across all models, which is kept consistent across all GPU architectures via gradient accumulation. We benchmark the video models with the following perceptual metrics: SSIM~\cite{hore2010image}, PSNR~\cite{hore2010image}, LPIPS~\cite{zhang2018unreasonable}, FVD~\cite{unterthiner2018towards}, and FID~\cite{heusel2017gans} using standard implementations. For FVD, we use features extracted by the R3D-18 model, which was pretrained on the Kinetics-400 dataset, available via torchvision. We report aggregate metrics computed on the stacked views for about $100$ trajectories per dataset subset. For each dataset, we follow standard procedures, with training splits for training and val/test splits used for evaluation. 
We provide additional details along with inference and timing results in~Appendix.~\ref{app:exp_setup} and Appendix.~\ref{app:timing_stats}.

\p{Baselines}
We benchmark \algname cross-embodiment video models against the state-of-the-art Ctrl-World~\cite{guo2026ctrlworldcontrollablegenerativeworld} baseline in the DROID environment. To ensure a fair comparison, we omit baselines with mismatched compute or inference requirements, including models that cannot perform autoregressive generation initialized from a single frame and those that require privileged information like robot URDFs and simulators. Although state-of-the-art baselines exist in the Bridge environment (e.g., WorldGym~\cite{quevedo2025worldgym}, Cosmos-Predict 2.5~\cite{nvidia2026worldsimulationvideofoundation}), we found these baselines to underperform our internal Bridge video models. Consequently, we train a new Bridge baseline (\texttt{Bridge-Base}) that outperforms existing alternatives (see~\Cref{sec:exp_cross_vs_single}). We train all baselines using the same model architecture and hyperparameters as \algname to control for confounding variables, noting the significant effect that hyperparameters typically have on performance. This experiment design enables us to isolate the impacts of different components of \algname on future-prediction fidelity. Comprehensive experiment results are provided in Appendix~\ref{app:real_world_experiments}.

\begin{figure*}[t]
    \centering
    \includegraphics[width=\textwidth]{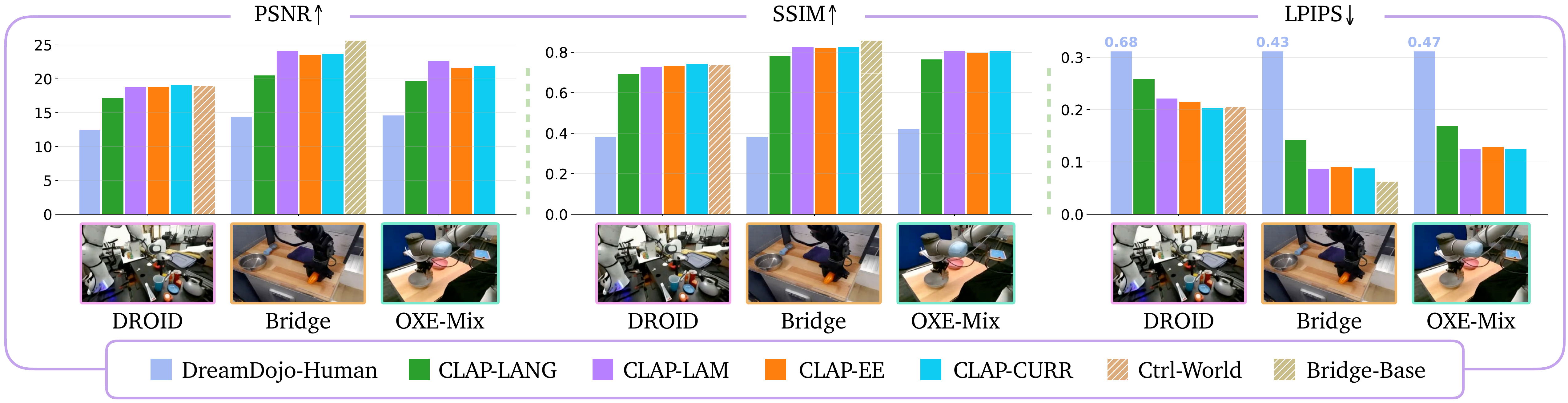}
    \caption{\textbf{Comparison to SOTA single-embodiment baselines.} \algname approaches or exceeds SOTA performance on the DROID platform, showing only minor fidelity degradation in simpler domains like Bridge. Across the diverse mix of embodiments (OXE-Mix), \alglam and \algcurr achieve the highest prediction accuracies, closely followed by \algee.}
    \label{fig:cross_vs_single_embodiment}
\end{figure*}

\begin{figure*}[t]
    \centering
    \includegraphics[width=\textwidth]{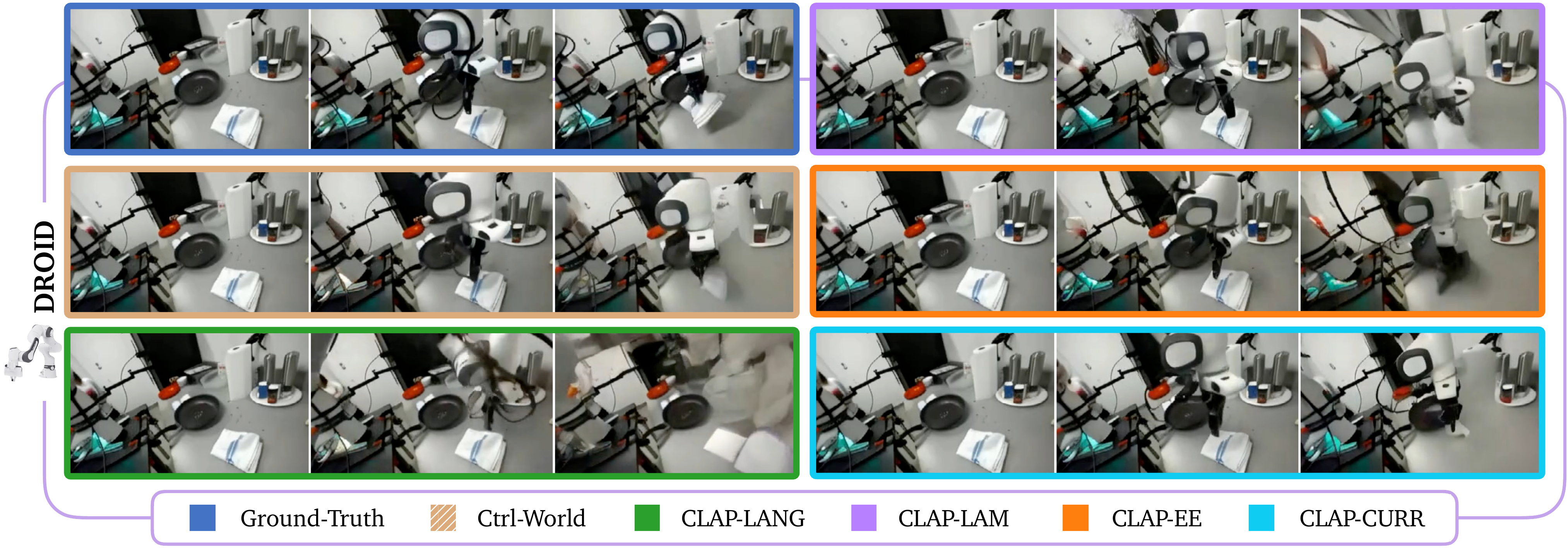}
    \caption{\textbf{\algname's performance on DROID.} \algcurr combines the strengths of \alglam and \algee to achieve high-fidelity dynamics prediction with fine-grained per-frame controllability, outperforming SOTA single-embodiment baselines. Meanwhile, \alglang delivers comparable performance over shorter horizons, though it is constrained by compounding errors over longer trajectories.}
    \label{fig:cross_vs_single_embodiment_droid}
\end{figure*}

\begin{figure*}[t]
    \centering
    \includegraphics[width=\textwidth]{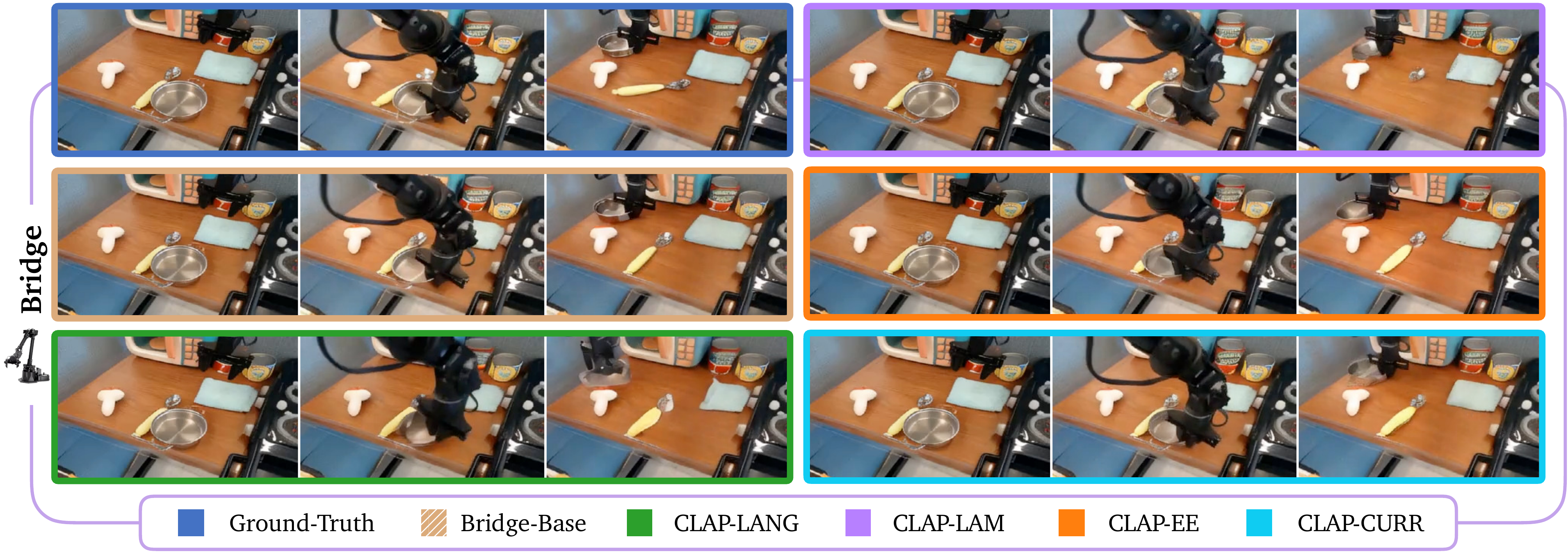}
    \caption{\textbf{\algname's performance on Bridge.} Relative to the SOTA baseline, \algname achieves high-accuracy future prediction on the Bridge platform; however, the performance of \alglang degrades over extended prediction horizons.}
    \label{fig:cross_vs_single_embodiment_bridge}
\end{figure*}

\begin{figure}[t]
    \centering
    \includegraphics[width=\columnwidth]{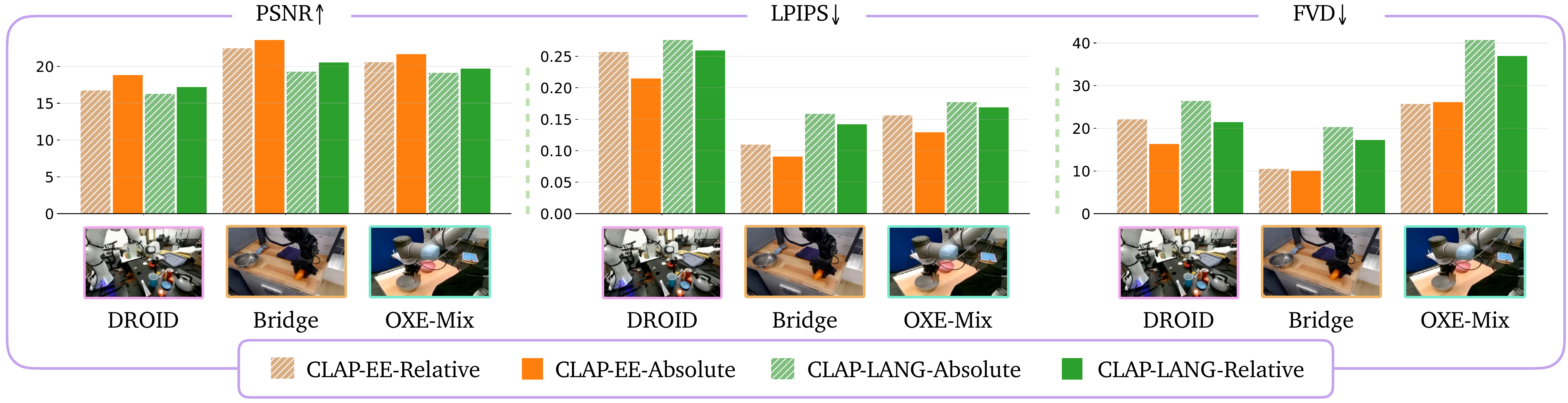}
    \caption{\textbf{Comparison between relative-action and absolute-action spaces.} Action-conditioned video generation in absolute-action spaces yields superior prediction fidelity; however, these gains are limited to future prediction tasks conditioned on end-effector-actions. In contrast, relative actions achieve higher prediction fidelity for language-action-conditioned video models.}
    \label{fig:relative_vs_absolute_actions}
\end{figure}

\subsection{Do cross-embodiment video models surpass their single-embodiment counterparts?}
\label{sec:exp_cross_vs_single}
We benchmark \algname's cross-embodiment video models against state-of-the-art single-embodiment baselines across diverse robot datasets, with a focus on dominant platforms like Bridge and DROID (\Cref{fig:cross_vs_single_embodiment}). We structure our evaluation along the following core axes:

\p{Action-space Harmonization}
We evaluate how different action-space harmonization schemes impact future-prediction fidelity. In doing so, we identify fine-grained trade-offs between end-effector and language action-spaces, revealing key differences in how well each integrates with diverse data sources. From~\Cref{fig:cross_vs_single_embodiment}, end-effector-action spaces enable fine-grained dynamics prediction across multiple robot morphologies, yielding superior prediction fidelity compared to language-action spaces. This divergence aligns with intuition: end-effector-action harmonization offers continuous precision, whereas coarse language actions suffer from the limitations of discrete tokenization. Consequently, end-effector-action-conditioned video world models resolve the granular action signals that are crucial for high-fidelity predictions. Notably, latent actions can match or surpass end-effector actions in future-prediction accuracy, defining the frontier in cross-embodiment video modeling across a diverse mix of robot platforms (OXE-Mix). Latent-action harmonization uniquely unlocks internet-scale training on unlabeled (action-free) video data (e.g., human video data) compared to other action harmonization schemes. However, these breakthroughs come at the expense of downstream deployment because physical robots operate in geometric action spaces, creating a mismatch between training and inference conditions. Addressing these limitations, \algcurr seamlessly combines the strengths of latent actions with those of end-effector actions to achieve high-fidelity cross-embodiment video models capable of zero-shot transfer to real-world deployment use-cases, without impeding data scaling. \Cref{fig:cross_vs_single_embodiment_droid,fig:cross_vs_single_embodiment_bridge} demonstrate these findings. On the more challenging DROID environment, \algcurr closely follows the ground-truth robot trajectory across all frames, improving upon the prediction fidelity of \alglam and \algee. On the Bridge platform, all \algname's models deliver high-accuracy future predictions. \alglang achieves comparable performance over shorter prediction horizons, but it is ultimately constrained by compounding errors over longer horizons.

\p{Human Videos vs. Multi-morphology Videos}
We investigate if human video data can function as a replacement for multi-morphology robot data in training cross-embodiment video models via the \texttt{DreamDojo-Human} baseline. We use the DreamDojo IDM trained exclusively on human videos to extract latent actions which serve as action-conditioning inputs for the video model. \Cref{fig:cross_vs_single_embodiment_droid} shows that robot video data is crucial for high-fidelity cross-embodiment video modeling, underscored by the significantly lower prediction accuracies achieved by \texttt{DreamDojo-Human}. For example, \algname's video models improve the \texttt{DreamDojo-Human}'s LPIPS score by at least $61\%$ on the DROID platform, with similar gains on all other perceptual metrics and robot environments. These findings underscore that although cross-embodiment video models can learn foundational physical priors from human video data, robot data is essential for effective transfer of these priors to robot morphologies, establishing it as crucial ingredient for training generalizable cross-embodiment video models.

\p{Cross-Embodiment vs. Single-Embodiment}
Single-embodiment video models have set the state-of-the-art (SOTA) in action-conditioned video generation; however, can cross-embodiment video models match their performance? Here, we compare \algname's cross-embodiment video models to the SOTA baselines. As shown in \Cref{fig:cross_vs_single_embodiment}, \algname's video models (e.g., \alglam, \algee)  match the performance of the SOTA baseline in the more challenging DROID environment. This finding challenges prevailing wisdom for two core reasons. First, \algname's cross-embodiment video models have the same number of model parameters and architecture as the single-embodiment video models. Given that the single-embodiment baselines are purpose-built-and-trained for future prediction on a specific robot platform, one would expect them to outperform \algname's cross-embodiment models. Second, to match the performance of single-embodiment baselines, cross-embodiment models typically require training on the union of all single-embodiment data alongside a significantly increased number of training steps. Surprisingly, \algname's cross-embodiment models match their single-embodiment counterparts despite utilizing effectively fewer DROID samples with the same training setup. 
These findings suggest that \algname learns generalizable physical priors from cross-embodiment data, enabling high-fidelity predictions even with limited domain-specific supervision.
This paradigm has the potential to transform how action-conditioned video models are trained, mirroring recent breakthroughs in multi-modal foundation models for vision and language generation.
In the simpler Bridge environment, \algname delivers high-accuracy dynamics predictions but does not precisely match our Bridge baseline (\texttt{Bridge-Base}), which is itself a new state-of-the-art baseline. (Prior video models achieve PSNR scores below 25 on the Bridge dataset.) The qualitative differences in the model's performance on the Bridge platform are mostly marginal, as shown in \Cref{fig:cross_vs_single_embodiment_bridge}.

\begin{figure}[t]
    \centering
    \includegraphics[width=\columnwidth]{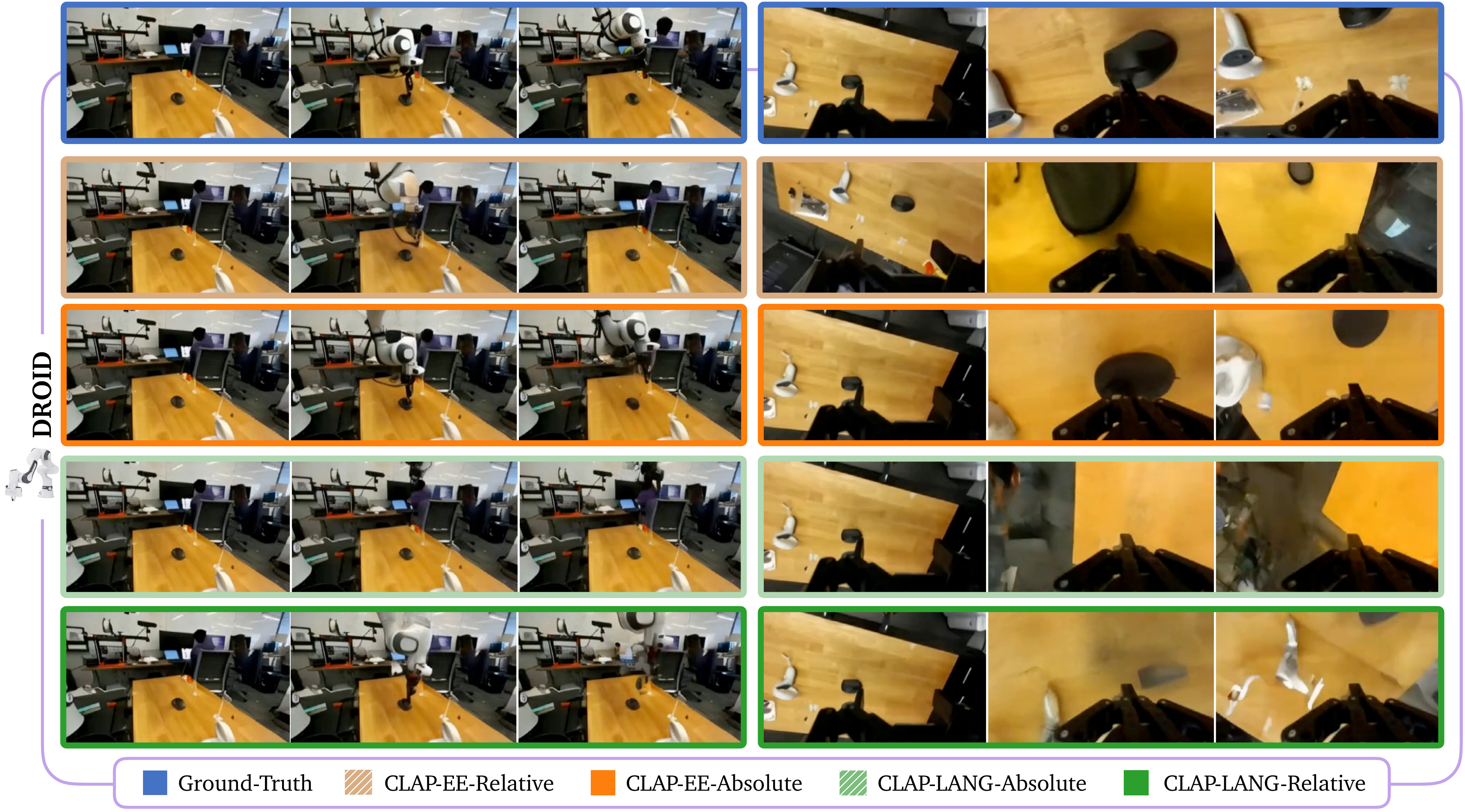}
    \caption{\textbf{Comparison between relative-action and absolute-action spaces in DROID.} In end-effector-action-conditioned video generation, absolute-action spaces enable fine-grained controllability over extended prediction horizons, an advantage that is not observed with relative-action spaces. Conversely, language-action-conditioned video models achieve higher-accuracy future prediction with relative actions.}
    \label{fig:relative_vs_absolute_actions_droid}
\end{figure}

\subsection{Do relative-action spaces outperform absolute-action spaces in cross-embodiment video modeling?}
\label{sec:exp_rel_vs_abs_action_spaces}
Recent work on video world models utilize relative-action spaces in action-conditioned video models, motivated by prevailing assumptions that relative actions offer a narrower action space with reduced modeling complexity. Our investigation of the impacts of relative-action and absolute-action spaces on cross-embodiment video modeling reveals nuanced findings, discussed in the following subsections:

\p{End-effector-action Conditioning}
From \Cref{fig:relative_vs_absolute_actions}, relative-action spaces underperform absolute-action spaces in future prediction conditioned on end-effector actions across all perceptual metrics and robot environments, e.g., by about $14.6\%$ in LPIPS in the DROID environment. This finding contradicts standard assumptions on the superiority of relative actions. Although relative actions might provide a less complex action space, they remain highly susceptible to compounding errors that degrade prediction fidelity, especially over extended horizons. In contrast, absolute-action spaces provide stronger robustness to compounding errors, yielding higher prediction accuracies, enabled by the action-space normalization strategies discussed in \Cref{sec:method_action_harmonization}. These findings are demonstrated in \Cref{fig:relative_vs_absolute_actions_droid}, where \texttt{CLAP-EE-Absolute} generates future frames that are more closely aligned with the ground-truth compared to \texttt{CLAP-EE-Relative}. These performance gains hold over both shorter and longer prediction horizons.

\p{Language-action Conditioning}
In language-action-conditioned video modeling, relative-action spaces achieve superior fidelity in dynamics prediction compared to absolute actions (see \Cref{fig:relative_vs_absolute_actions}). These findings likely stem from the limited resolution of discrete tokenization spaces. Under these constraints, the compact bounds of relative actions utilize the tokenization space more effectively, yielding higher prediction accuracies. In contrast, the wider input distribution of absolute-action spaces limits prediction fidelity due to these bottlenecks. \Cref{fig:relative_vs_absolute_actions_droid} visualizes these challenges. Unlike \texttt{CLAP-LANG-Absolute}, \texttt{CLAP-LANG-Relative} yields higher-fidelity predictions that approximately match the ground-truth.

\begin{figure}[t]
    \centering
    \includegraphics[width=\columnwidth]{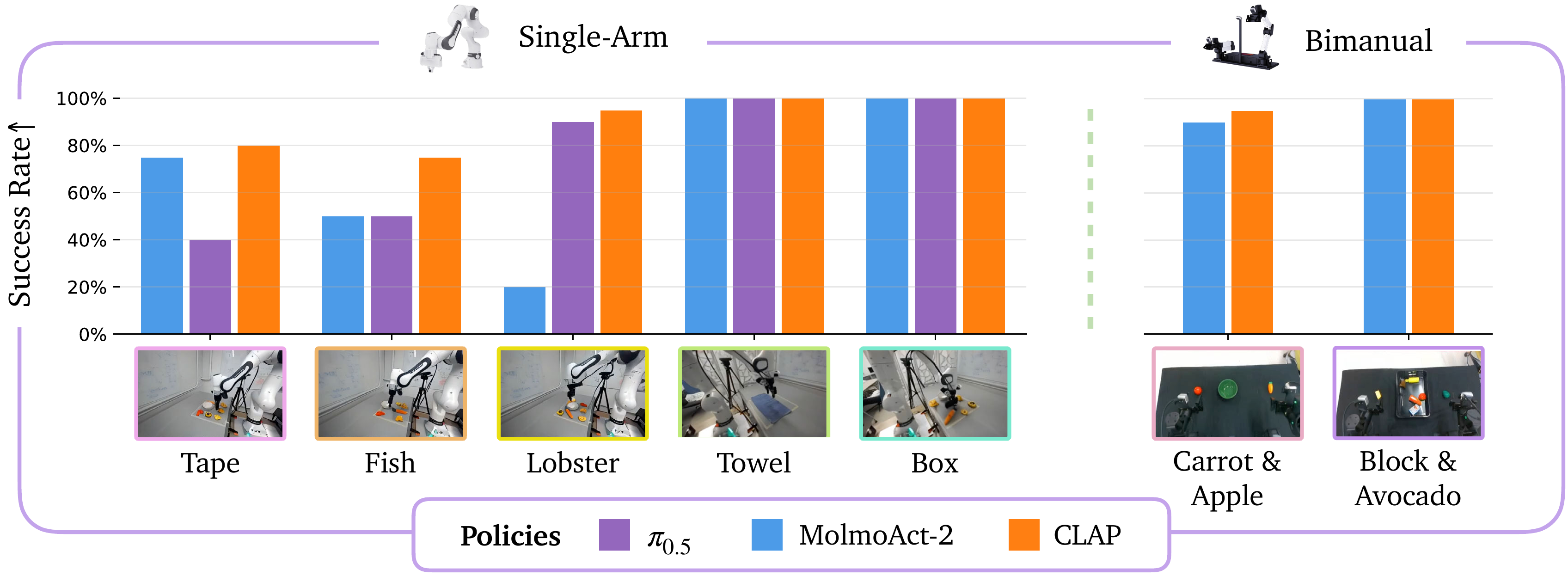}
    \caption{\textbf{Inference-time cross-policy planning with \algname.} In contrast with prior work, \algname optimizes over action proposals from multiple SOTA policies for robust robot manipulation across diverse tasks, boosting the success rates of $\pi_{0.5}$ and MolmoAct-2. Moreover, we stress-test \algname in bimanual robot manipulation beyond its standard operational domain, demonstrating improvements in success rates on a bimanual YAM robot.}
    \label{fig:inf_time_planning_metrics}
\end{figure}

\begin{figure}[t]
    \centering
    \includegraphics[width=\columnwidth]{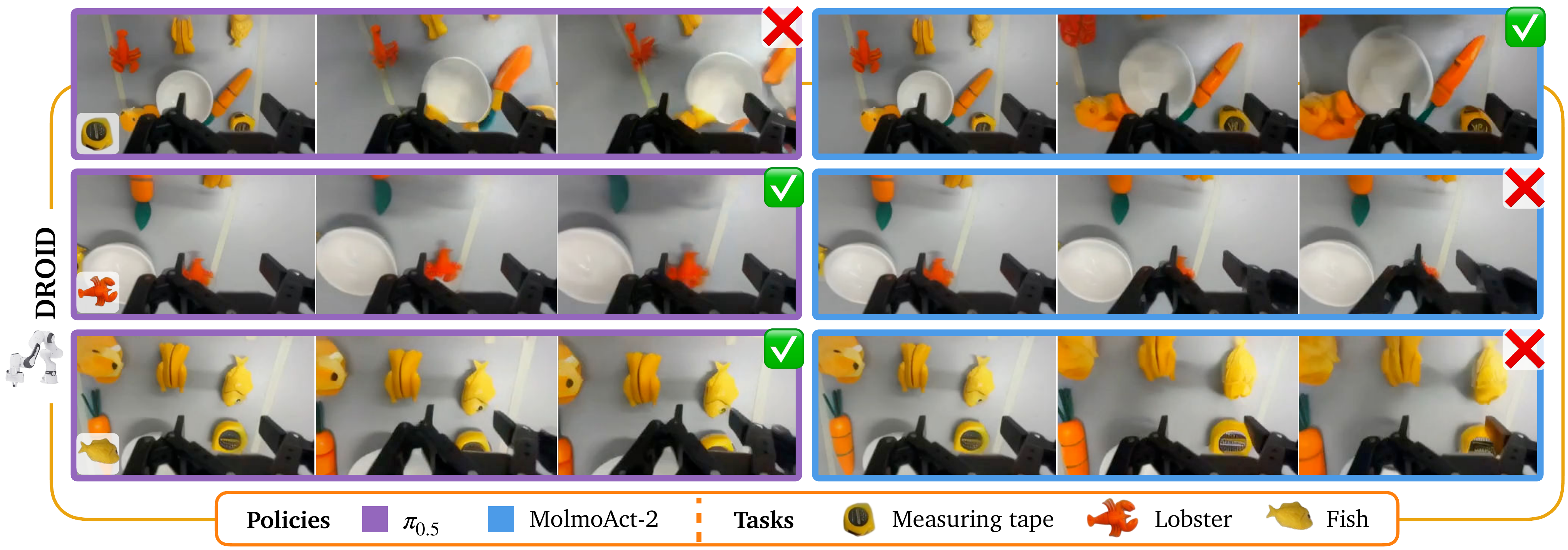}
    \caption{\textbf{Inference-time cross-policy planning on single-arm Franka robot (DROID).} \algname predicts future frames conditioned on robot action proposals generated by multiple policies to maximize task-aligned rewards. (First row)~In the \emph{tape} task, \algname identifies potential semantic confusion in $\pi_{0.5}$ and thus selects the more consistent action proposals from MolmoAct-2 for execution. (Second row)~In the \emph{lobster} task, \algname detects a failure in MolmoAct-2's semantic reasoning and thus executes actions from $\pi_{0.5}$. (Third row)~In the \emph{fish} task, \algname selects the more promising action from $\pi_{0.5}$, although both policies' action proposals are likely to succeed based on \algname's predictions. (The first frame per block is the initial frame; all other frames are future frames generated by \algname.)}
    \label{fig:inf_time_planning_viz_single_arm}
\end{figure}

\begin{figure}[t]
    \centering
    \includegraphics[width=\columnwidth]{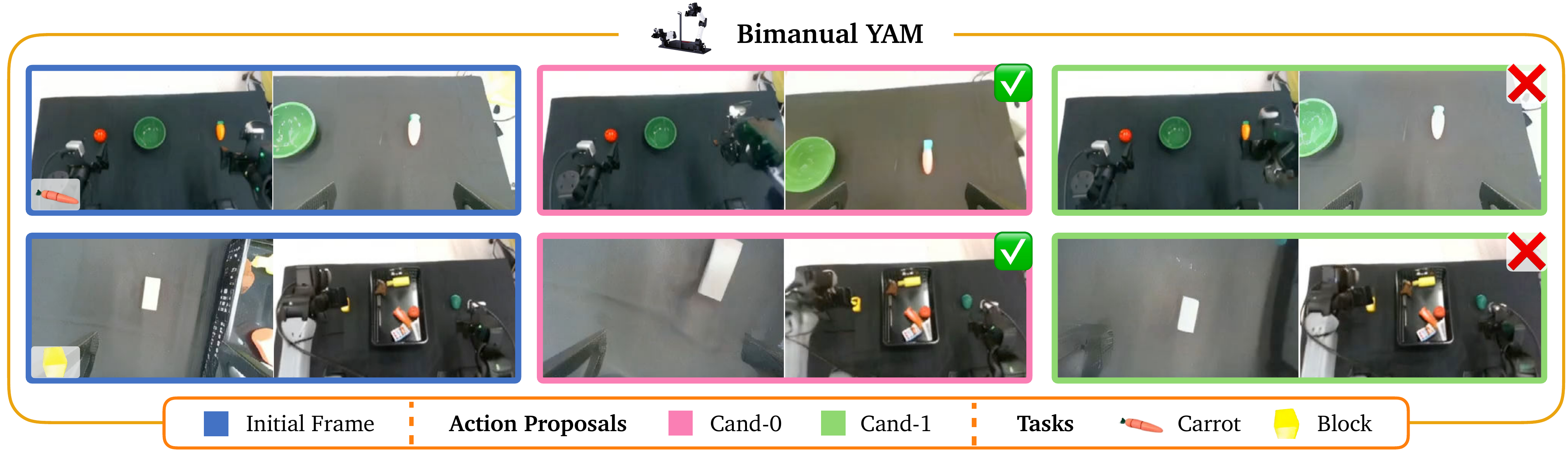}
    \caption{\textbf{Inference-time cross-policy planning on bimanual YAM.} Although \algname was only trained on single-arm robot data, \algname generates sufficiently accurate future frames conditioned on candidate robot actions for planning with bimanual robots. (First row)~\algname selects the candidate action that brings the right arm closer to the target object (\emph{carrot}). (Second row)~\algname selects the action that makes greater progress towards the target object (\emph{block}).}
    \label{fig:inf_time_planning_viz_bimanual}
\end{figure}

\subsection{Do cross-embodiment video models achieve zero-shot real-world generalization?}
\label{sec:exp_real_world}
We investigate the ability of \algname's cross-embodiment models to generalize to real-world robot manipulation tasks without any additional data (i.e., in zero-shot). Prior work~\citep{kim2026cosmospolicyfinetuningvideo, mei2026inferencetime} has demonstrated that single-embodiment video models are effective in test-time planning in robot manipulation tasks given a set of action proposals. However, generating a set of informative, yet diverse action proposals remains difficult. This challenge has spurred the design of heuristic-based strategies that perturb nominal actions from a single policy with random sampling and noise augmentation. However, these strategies suffer from lower execution quality and require high sample complexity for adequate coverage. To address these limitations, we deploy \algname as a world modeling backbone in inference-time planning across multiple robot policies, generating a set of diverse candidate actions that typically contains optimal trajectories. We use \alglam as the base video model with a lightweight adapter to map end-effector actions to latent actions. This experiment design enables us to assess the lower bounds of performance among \algname's non-language-action models, given the more significant mismatch between end-effector and latent actions.
We generate candidate actions from the SOTA policies $\pi_{0.5}$~\citep{intelligence2025pi} and MolmoAct-2~\citep{fang2026molmoact2actionreasoningmodels} across five tasks on the single-arm Franka Panda robot in a DROID environment. To examine the robustness of \algname's models beyond their normal operational boundaries (i.e., in single-arm robot manipulation), we stress-test them in bimanual robot manipulation. We discuss these results in the following subsections and provide additional experiments in finetuning robot manipulation policies with video-model-based reinforcement learning in~Appendix~\ref{app:real_world_experiments}.

\p{Single-arm Manipulation}
We assess the success rate of $\pi_{0.5}$ and MolmoAct-2 across five tasks, spanning pick-and-place --- with target objects \emph{tape}, \emph{fish}, \emph{box}, \emph{lobster} --- and a towel-folding task. At inference time, \algname's planner predicts future frames conditioned on $k$ sampled actions from each policy and scores each future trajectory using a VLM given the initial and final frames (consisting of left-camera and wrist-camera views). We set ${k = 2}$ with GPT-5 mini as the VLM, given its superior performance on the RoboRewardBench evaluation benchmark~\cite{lee2026roboreward}. Thereafter, the robot executes the trajectory with the maximum predicted reward. \Cref{fig:inf_time_planning_metrics} reports the success rates of \algname's planner compared to the baseline policies across $20$ trials per policy on each task, demonstrating that \algname matches or improves the base success rates of all policies across all tasks. Notably, the base performance of each policy varies significantly with the target object. This variation can be attributed to the brittle semantic reasoning capabilities of VLAs, especially in tasks involving relatively uncommon target objects. \algname addresses this core limitation by combining the strengths of multiple policies to compensate for their individual weaknesses. This phenomenon is evident in the \emph{tape}, \emph{fish}, and \emph{lobster} tasks. While MolmoAct-2 performs well on the \emph{tape} task, it struggles on the \emph{lobster} task; in contrast, $\pi_{0.5}$ performs well on the \emph{lobster} task, but struggles on the \emph{tape} task. Notably, even when both policies struggle on a common task, \algname still achieves higher success rates, because planning is executed sequentially in a receding-horizon fashion, which ultimately improves robustness.
\Cref{fig:inf_time_planning_viz_single_arm} shows the predicted future trajectories generated by \algname conditioned on candidate actions from both policies. The first frame per block is the initial frame; all other frames are future frames generated by \algname. In the \emph{tape} task~(first row), \algname detects potential semantic confusion in $\pi_{0.5}$ based on its sampled actions, and selects the actions proposed by MolmoAct-2 to maximize task-aligned rewards. Conversely, in the \emph{lobster} task, \algname identifies potential semantic failure in MolmoAct-2 but stronger task-alignment in $\pi_{0.5}$'s proposed actions. In the \emph{fish} task, \algname selects $\pi_{0.5}$ actions, which are more promising since they make better progress towards completing task.

\p{Bimanual Manipulation}
We evaluate the success rate of \algname in inference-time planning with the bimanual YAM robot using MolmoAct-2 bimanual-YAM as the base policy. We consider two multi-stage tasks: ``put the \textless object A\textgreater\ in the bowl/bin, then put the \textless object B\textgreater\ in the bowl/bin,'' with the following object pairs: \{\emph{carrot, apple}\}, \{\emph{block, avocado}\}. We alternate the order of each object per pair to assess both semantic understanding and the manipulation capability of the base policy. 
To address the mismatch in the action spaces of the $14$-dimensional bimanual robot and $7$-dimensional \algname's inputs, we dynamically select between the two arms, prioritizing the arm whose action chunk has a greater magnitude. The video model uses this information, along with other temporal and semantic cues, to resolve robot-arm motion in the generated videos.
\Cref{fig:inf_time_planning_metrics} summarizes the success rates of each method with $10$ trials per task. Despite the gap between the \algname's normal operational domain and the test-time conditions, \algname still achieves higher success rates compared to the base policy, with the only failure occurring when the planner dropped the second target object next to, but outside the bowl. From~\Cref{fig:inf_time_planning_viz_bimanual}, \algname generates future predictions accurate enough to distinguish promising action proposals from lower-quality candidates, driving its superior performance. We discuss model adaptation strategies for fine-grained bimanual manipulation in~\Cref{sec:exp_few_shot_adaptation}.

\begin{figure}[t]
    \centering
    \includegraphics[width=\columnwidth]{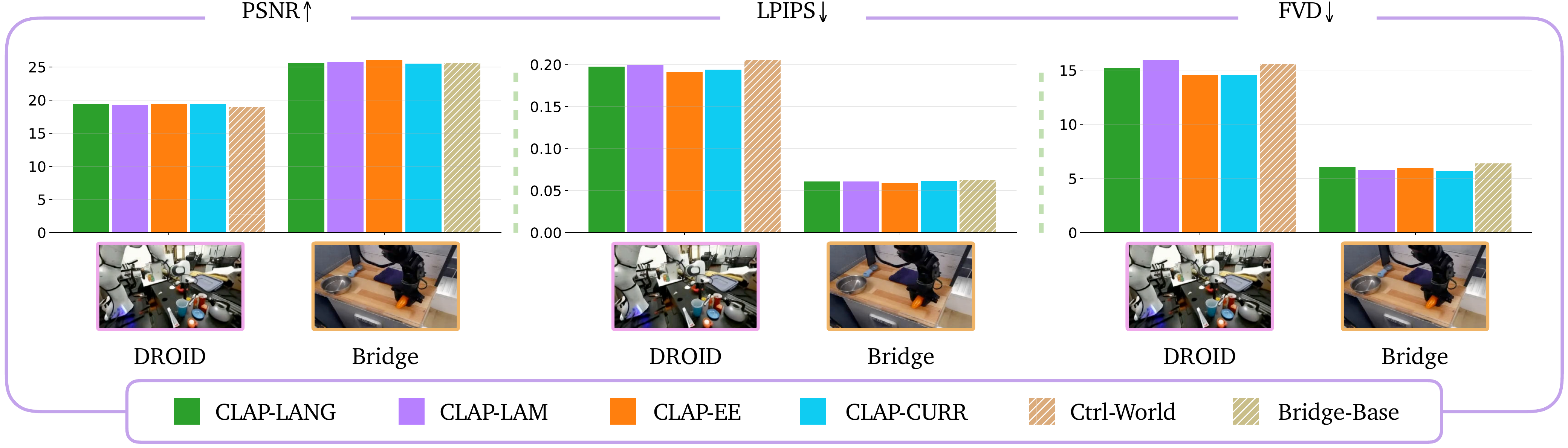}
    \caption{\textbf{Data-efficient adaptation to pretrained embodiments.} \algname's cross-embodiment models provide a crucial foundation for data-efficient adaptation to pretrained embodiments. With \algname, all post-trained models surpass SOTA baselines on nearly all perceptual metrics.}
    \label{fig:beyond_zero_shot_pretrained_embodiments}
\end{figure}

\begin{figure}[t]
    \centering
    \includegraphics[width=\columnwidth]{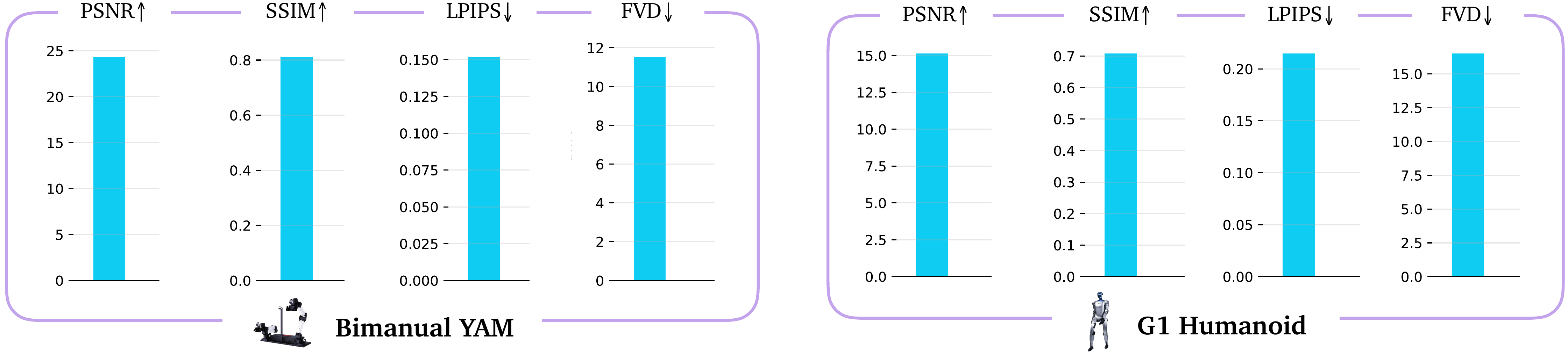}
    \caption{\textbf{Few-shot adaptation to novel embodiments (bimanual YAM and G1~humanoid).} \algname facilitates few-shot adaptation of cross-embodiment models to the $14$-dimensional and $26$-dimensional action spaces of the bimanual YAM robot and G1 humanoid, respectively, for high-fidelity future prediction.}
    \label{fig:beyond_zero_shot_novel_embodiments_metrics}
\end{figure}

\begin{figure}[t]
    \centering
    \includegraphics[width=\columnwidth]{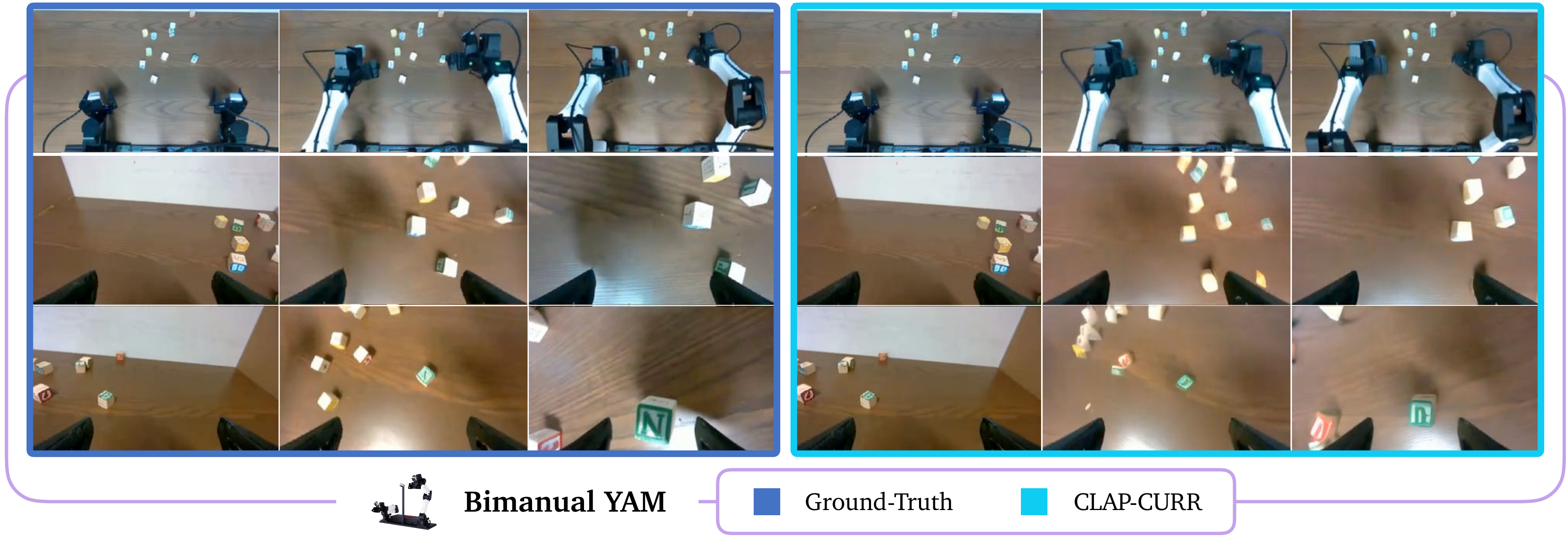}
    \caption{\textbf{Few-shot adaptation to bimanual YAM.} After few-shot finetuning, \algname achieves high-fidelity future prediction on the bimanual YAM robot and accurately resolves each arm's motion spatially and temporally.}
    \label{fig:beyond_zero_shot_novel_bimanual_yam_viz}
\end{figure}

\begin{figure}[t]
    \centering
    \includegraphics[width=0.85\columnwidth]{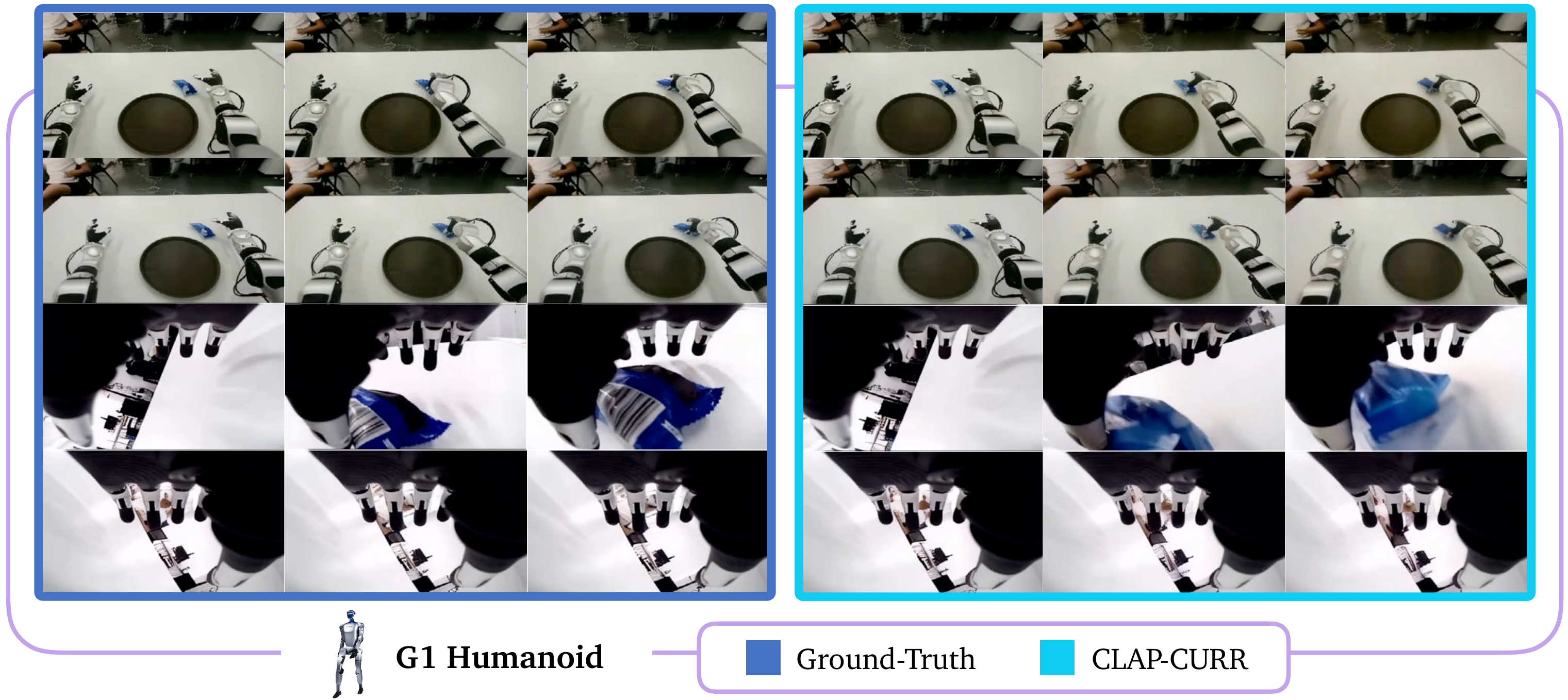}
    \caption{\textbf{Few-shot adaptation to G1 humanoid.} Using \algcurr as the backbone, the finetuned video model achieves high-accuracy dynamics prediction on the G1 humanoid.}
    \label{fig:beyond_zero_shot_novel_humanoid_viz}
\end{figure}

\subsection{Beyond Zero-Shot: Does \algname facilitate data-efficient adaptation to target embodiments?}
\label{sec:exp_few_shot_adaptation}
We evaluate the effectiveness of \algname as a foundation for few-shot adaptation of video world models to target embodiments. We ground our evaluations along two primary axes: pretrained embodiments and novel embodiments. In each setting, we drop incompatible components of the action head to match the native end-effector action space of the target embodiment but retain all other parameters.

\p{Pretrained Embodiments}
We finetune \algname's cross-embodiment models on the Bridge and DROID environments and benchmark their performance against SOTA single-embodiment baselines to assess the ability of the finetuned models to leverage physical priors from the cross-embodiment models for higher-fidelity dynamics prediction. \Cref{fig:beyond_zero_shot_pretrained_embodiments} supports this hypothesis. Notably, all post-trained models surpass the SOTA baselines across nearly all perceptual metrics, suggesting a successful transfer of physical priors from the base models. In the DROID environment, finetuning bridges the gap between \alglang and other methods, highlighting the benefits of higher-resolution action-conditioning schemes for prediction fidelity. Similarly, in the Bridge environment, finetuning enables all cross-embodiment models to close the gap with and ultimately outperform the single-embodiment baseline.
These findings establish a novel paradigm for training single-embodiment video world models. Concretely, our experiments show that finetuning cross-embodiment video models into single-embodiment variants is a more promising path towards high-fidelity dynamics prediction than the prevailing paradigm of training them from scratch or from less-aligned backbones like SVD~\cite{blattmann2023stable} or WAN~\cite{wan2025}.

\p{Novel Embodiments}
Motivated by the generalizable physical priors encoded in cross-embodiment video models, we investigate their few-shot adaptation to novel embodiments, including bimanual robots and humanoids. Following the same finetuning framework, we adapt \algcurr on the $14$-dimensional action space of the bimanual YAM robot ($7$-DoF per arm) and the $26$-dimensional action space of the G1 humanoid ($7$-DoF per arm and $6$-DoF per hand).
\Cref{fig:beyond_zero_shot_novel_embodiments_metrics} summarizes the performance of each model. The bimanual YAM video model achieves significantly higher perceptual scores because the YAM dataset features less challenging, characterized by slower motion and smoother frame-to-frame transitions, compared to the more visually demanding G1 humanoid environment. In~\Cref{fig:beyond_zero_shot_novel_bimanual_yam_viz}, the bimanual YAM video model accurately predicts the motion of each arm spatially and temporally, producing generated frames that closely match the ground-truth. Similarly,~\Cref{fig:beyond_zero_shot_novel_humanoid_viz} demonstrates that the G1 humanoid video model achieves high-fidelity future prediction of both arm and hand movements. These results underscore the effectiveness of \algname as a foundational backbone for few-shot adaptation to novel embodiments.

\section{Conclusion, Limitations, and Future Work}
\label{sec:conclusion}
We present \algname, a cross-embodiment action-conditioned video modeling framework that learns generalizable physical priors from heterogeneous human and robot videos. \algname bridges the disparate action spaces of diverse embodiments with end-effector poses, language instructions, and learned latent action representations. To address the inherent limitations of these action representations, \algname introduces a curriculum-based recipe that combines the strengths of latent actions  and end-effector actions for data scaling with unlabeled video data while achieving zero action-space mismatch in real-world deployments. These core contributions yield cross-embodiment video models that approach or surpass state-of-the-art single-embodiment baselines. Further, \algname establishes a novel framework for training single-embodiment action-conditioned video models through sample-efficient adaptation of cross-embodiment models. Crucially, we demonstrate \algname's zero-shot generalization to real-world tasks in inference-time cross-policy planning and reinforcement-learning-based policy finetuning in video world models.

Despite its high prediction fidelity, \algname is still prone to hallucinations, like other video world models. Mitigating these hallucinations is critical for trustworthy integration in diverse robotics applications, such as planning, policy evaluation, and policy finetuning, presenting an exciting direction for future work on hallucination detection and mitigation, e.g., via uncertainty quantification~\citep{mei2026worldmodelsknowdont, mei2025confidentvideomodelsempowering}. Additionally, \algname is primarily trained on single-embodiment robot data and human video data, extending this pipeline to bimanual and humanoid robot data would further scale the training data diversity, which could ultimately boost future-prediction accuracy and zero-shot generalization, constituting a promising path forward. Moreover, greater diversity in the training data could improve \algname's adaptation efficiency to novel embodiments. While model and data scaling provide immediate pathways for continuous improvement, they incur significant training and inference costs. Consequently, future research into efficient model architectures will be critical.

Ultimately, our work establishes cross-embodiment learning as a catalyst for breakthroughs in video world modeling, mirroring the paradigm shifts observed in large language models.

\section*{Safety, Data Privacy, and Consent}
This paper presents work advancing the foundations of artificial intelligence (AI) in robotics. Because this work is primarily computational and theoretical, it does not involve human participants, live user studies, or personally identifiable information (PII); consequently, Institutional Review Board (IRB) approval was not required. All experiments were conducted using publicly available benchmarks and open-source datasets in accordance with their respective licenses. While broader applications of AI in robotics carry potential safety and societal implications, we do not foresee any direct, malicious, or high-risk negative societal consequences uniquely tied to this fundamental algorithmic research.

\section*{Acknowledgments}
The authors would like to thank Prof. Anirudha Majumdar, Mingtong Zhang, and members of the Princeton IRoM lab for their patience, insightful discussions, and support.

\bibliographystyle{unsrtnat}
\bibliography{references.bib}

\clearpage

\beginappendix{
    \section*{Table of Contents}
    \startcontents[appendices] %
    \printcontents[appendices]{}{1}{} %
    \clearpage
    
    \section{Preliminaries}
\label{sec:app_prelims}
Video world models predict future outcomes conditioned on per-frame actions, starting from an initial camera observation. Furthermore, to capture temporal context such as robot velocities that dictate future evolution, video world models are often conditioned on a history of past observations. 

In our work, the video world model $\mcal{V}_{\theta}$ takes as input a language instruction $l$, a sequence of observations spanning past history and the current frame ${\{o_{\tau}\}_{\tau = t - H}^{t}}$, and a sequence of future per-frame actions ${\{a_{\tau}\}_{\tau = t}^{t + P}}$. From these inputs, $\mcal{V}_{\theta}$ predicts the resulting sequence of future frames ${\{o_{\tau}\}_{\tau = t + 1 }^{t + P}}$ via:
\begin{equation}
    \{o_{\tau}\}_{\tau = t + 1 }^{t + P} \sim \mcal{V}_{\theta}(\cdot \mid l, \{o_{\tau}\}_{\tau = t - H}^{t}, \{a_{\tau}\}_{\tau = t}^{t + P}),
\end{equation}
where $H$ and $P$ denote the history and prediction horizons, respectively.

\section{Method}
\label{app:method}

\subsection{Action Representations}
\label{app:action_representations}
\p{End-effector Actions}
Each demonstration is stored as a trajectory of end-effector states aligned to the video frames. A single state is a $7$-dimensional vector
${s_t = \big[\,x_t,\; y_t,\; z_t,\; \phi_t,\; \theta_t,\; \psi_t,\; g_t\,\big],}$
where $(x_t, y_t, z_t)$ is the gripper position in meters in the robot base frame, $(\phi_t, \theta_t, \psi_t)$ are the roll--pitch--yaw orientation Euler angles in radians, and $g_t$ is the raw gripper signal. Depending on the evaluation setup, end-effector actions are computed either in absolute coordinates or relative to the initial frame of the history window.
\algee directly encodes $s_{t}$ into a 1024-dimensional vector which is passed into the video model as a conditioning signal.

\p{Language Actions.}
\alglang maps per-frame end-effector actions to natural language, producing a single natural-language string per timestep via the following template:

\begin{tcolorbox}[tbox_style, title=Language Action Format]
x=$\langle \rangle$, y=$\langle \rangle$, z=$\langle \rangle$, roll=$\langle \rangle$,
pitch=$\langle \rangle$, yaw=$\langle \rangle$, gripper $\langle g\rangle$
\end{tcolorbox}
We express the gripper position numerically, normalized between zero and one. 
\Cref{tab:gripper} summarizes how we compute the normalized gripper position $g_{t}$ from the provided gripper signal $\bar{g}_{t}$ in each dataset.
Depending on the action-conditioning space, language actions are expressed either in absolute coordinates or relative to the current frame.

\p{Latent Actions}
 We infer latent actions directly from the observed video. The latent action model (see Appendix~\ref{app:model_architecture}) encodes the transition between consecutive frames into
  a compact continuous latent vector, yielding a trajectory of $T$ latent actions
  $z_t \in \mathbb{R}^{32}$, stored as a per-episode array aligned with the video
  frames. Because these actions are learned purely from observation, they are
  embodiment- and sensor-agnostic, providing a uniform action signal that is
  defined even for datasets whose proprioceptive action labels are noisy,
  inconsistent, or unavailable.

\subsection{Model Architecture}
\label{app:model_architecture}

\p{Latent Action Model}
We model latent actions with a variational autoencoder over consecutive video frames, following the latent-action formulation of Genie~\citep{bruce2024genie}.
Each training example is a clip of $T=2$ RGB frames at $240\times320$ resolution, with pixel values normalized to $[0,1]$. Frames are split into non-overlapping $16\times16$ patches, yielding a $15\times20=300$-token grid per frame, where each token is the flattened patch of dimension $3\cdot16^2 = 768$. We denote the resulting patch tensor $\mathbf{P}\in\mathbb{R}^{B\times T\times N\times d_p}$ with $N=300$ and $d_p=768$. The flattened patch is fed into the encoder, which is a factorized spatio-temporal transformer of $L_\text{enc}=24$ blocks operating at model width $d=1024$ with $16$ attention heads. The encoder produces a $d$-dimensional action-prompt token at the future frame(s). %

The model is trained with a standard $\beta$-VAE objective, combining pixel reconstruction and a KL prior-matching term:
\begin{equation}
\mathcal{L} = \underbrace{\big\|x_{t+1}-\hat{x}_{t+1}\big\|_2^2}_{\text{MSE reconstruction}} \;+\; \beta\cdot D_\text{KL}\!\big(\mathcal{N}(\boldsymbol{\mu}_t,\boldsymbol{\sigma}^2_t)\,\|\,\mathcal{N}(0,I)\big),
\end{equation}
with a small $\beta=10^{-6}$, which prioritizes reconstruction fidelity while still regularizing the latent space toward a unit Gaussian.

We optimize with AdamW (learning rate $\text{lr}=5\times10^{-6}$, weight decay $10^{-2}$, gradient clipping at $0.3$) under mixed-precision (fp16) on 8 GPUs with the distributed data parallel framework, at a global batch size of 64. The model is fine-tuned from a checkpoint pre-trained for 400K steps. As reported in~\Cref{tab:lam_data_mixture}, the training data is a weighted mixture of 10 robot- and egocentric-manipulation video datasets sampled with dataset-specific mixture weights and per-dataset frame-skip and multi-view stacking settings to normalize frame rate and camera layout across sources.

\begin{table}[t]
  \centering
  \small
  \caption{Per-dataset gripper normalization to a common gripper position signal
  $g_t \in [0,1]$. The width scheme uses the maximum gripper width
  $w_{\max} = 0.08$\,m.}
  \label{tab:gripper}
  \begin{tabular}{@{}llll@{}}
    \toprule
    Scheme & Raw signal $\bar{g}_{t}$ & Gripper position $g_t$ & Datasets \\
    \midrule
    Normal  & normalized $[0,1]$, high $=$ open    & $g_t = \bar{g}_t$ & bridge, fractal, droid \\
    Flipped & normalized $[0,1]$, high $=$ closed   & $g_t = 1 - \bar{g}_t$ & bc\_z, fmb \\
    Width   & gripper width (m), wide $=$ open      & $g_t = \bar{g}_t / w_{\max}$ & \makecell[l]{furniture\_bench, taco\_play,\\ austin\_sailor, stanford\_hydra,\\ utaustin\_mutex} \\
    Command & signed open/close events $\{-1,0,+1\}$ & \makecell[l]{integrate from open:\\ $+1 \rightarrow$ close, $-1 \rightarrow$ open} & berkeley\_autolab\_ur5 \\
    \bottomrule
  \end{tabular}
\end{table}

\p{Video Model}
\algname uses the SVD spatio-temporal UNet as its video backbone, interleaving spatial and temporal attention across four resolution stages with cross-attention dimension $1024$ and input channel dimension of $8$: four for the noised target latents and four for a per-frame conditioning latent. 
The frozen VAE encoder $\mathcal{E}$ maps each RGB frame to a latent $z_t \in \mathbb{R}^{4 \times H/8 \times W/8}$, and all generation occurs in this latent space. We split a window of $T$ frames into ${T_h = 6}$ history frames and ${T_f = 5}$ future frames, and train the model to denoise the future latents given the history $z_{1:T_h}$, the current observation $z_{T_h}$, and actions $a_{1:T}$.
We use EDM noise sampling~\citep{karras2022elucidatingdesignspacediffusionbased}, with $\log\sigma \sim \mathcal{N}(P_\text{mean}, P_\text{std}^2)$, $P_\text{mean}{=}0.7$, $P_\text{std}{=}1.6$.

A single interface maps any per-frame control signal to the $1024$-dimensional token space of the UNet cross-attention layers. End-effector and latent actions are encoded by a three-layer MLP with SiLU activations (with input dimensions $7$ and $32$ for end-effector actions and latent actions, respectively); per-frame language actions are encoded by a frozen CLIP text encoder. An optional task-level instruction embedding is added to the per-frame tokens, when available in the dataset. Critically, conditioning is applied at the \emph{frame level}: each generated frame attends to its own action token rather than a shared global token, which is what makes the model controllable. The action context is dropped with probability $5\%$ during training to enable classifier-free guidance.

During training, the VAE and CLIP encoder stay frozen; only the UNet is trained, using AdamW at learning rate $10^{-5}$, gradient clipping $1.0$, and mixed precision. At inference, we use the EDM sampler with $50$ steps, frame-wise classifier-free guidance, and autoregressive chunked rollout.

\begin{table}[t]
  \centering
    \begin{minipage}[t]{0.48\textwidth}
        \centering
        \caption{Training data mixture and per-dataset sampling weights for LAM.}
        \label{tab:lam_data_mixture}
        \begin{tabular}{l r}
        \toprule
        Training Dataset & Ratio \\
        \midrule
        EgoDex~\citep{hoque2026egodex}            & 30.0\% \\
        Bridge~\citep{o2024open}            & 14.8\% \\
        Fractal~\citep{o2024open}          & 14.2\% \\
        DROID~\citep{o2024open}              & 11.1\% \\
        BC-Z~\citep{o2024open}                 &  7.5\% \\
        FMB~\citep{o2024open}                  &  7.1\% \\
        Language Table~\citep{o2024open} &  4.4\% \\
        Taco Play~\citep{o2024open}       &  3.0\% \\
        Furniture Bench~\citep{o2024open}&  2.4\% \\
        RoboTurk~\citep{o2024open}        &  2.3\% \\
        \bottomrule
        \end{tabular}
    \end{minipage}
    \hfill
    \begin{minipage}[t]{0.48\textwidth}
        \centering
        \caption{Training data mixture and per-dataset sampling weights for the video models.}
        \label{tab:video_model_data_mixture}
        \begin{tabular}{l r}
        \toprule
        Training Dataset & Ratio \\
        \midrule
        EgoDex~\citep{hoque2026egodex}            & 2.5\% \\
        Bridge~\citep{o2024open}            & 15.0\% \\
        Fractal~\citep{o2024open}          & 1.50\% \\
        DROID~\citep{o2024open}              & 75.0\% \\
        BC-Z~\citep{o2024open}                 &  1.50\% \\
        FMB~\citep{o2024open}                  &  1.50\% \\
        Taco Play~\citep{o2024open}       &  1.50\% \\
        Furniture Bench~\citep{o2024open}&  1.50\% \\
        \bottomrule
        \end{tabular}
    \end{minipage}
\end{table}

\section{Experiments}
\subsection{Evaluation Setup}
\label{app:exp_setup}
\algname trains cross-embodiment video models on diverse datasets using the sampling ratios in~\Cref{tab:video_model_data_mixture}, proportional to the complexity and quaity of the individual datasets.c 
Multi-view camera images are vertically stacked into a single fixed-size frame (${576 \times 320}$) after resizing individual frames to ${192 \times 320}$. For uniformity, we repeat single-view camera inputs in datasets without multi-view images, although preliminary experiments revealed that such repetition is unnecessary.
In real-world experiments that directly utilize \alglam, we train a lightweight adapter that maps end-effector actions into the $32$-dimensional latent-action space, which serves as conditioning inputs for the video model. Its output head is zero-initialized; while the adapter is trained, the world model is kept frozen.
All models are trained on the Open X-Embodiment (OXE) datasets, with \alglam also incorporating the EgoDex dataset. The OXE-Mix training split includes the Bridge, DROID, FMB, Furniture Bench, and Taco Play datasets. We evaluate the trained models on the held-out test or validation splits with about $100$ trajectories each, and additionally report their average performance across the Bridge, DROID, and Taco Play datasets. Real-world table-top manipulation experiments are performed on the Franka Emika robot in the DROID configuration with two side cameras and a wrist camera and a Bimanual YAM robot setup with a top-view camera and a wrist-camera view per arm.

\subsection{Inference and Timing Results}
\label{app:timing_stats}
\algname requires under $12$~GB of VRAM at inference and can comfortably fit on consumer hardware like the RTX 3060, with each inference call consuming approximately $9.7$~GB (verified on the A100 and H200 GPU nodes). All timing measurements are evaluated with a nominal prediction of $11$ total frames and $25$ denoising steps, measured across 20 trials. At initialization, the timing for the first inference call depends on the status of the cuDNN kernel-autotuning cache: if a previous process has already initialized the kernel, the first call takes about $3$~seconds, but takes roughly $15$~seconds otherwise. Subsequent inference times vary by GPU architecture, averaging ${3.24\text{s} \pm 0.02\text{s}}$ on the A100-PCIe-40GB, ${2.88\text{s} \pm 0.00\text{s}}$ on the A100-SXM4-80GB, and ${1.49\text{s} \pm 0.00\text{s}}$ on the H200.

\subsection{Additional Results on Latent Action Models}
\label{app:latent_action_idm}

We train latent action models (LAMs) on our cross-embodiment dataset and individual single-embodiment subdatasets (Bridge and DROID) and evaluate the consistency of the proxy actions relative to the ground-truth trajectory. Using LAMs as inverse-dynamics models (IDMs), we extract latent actions for unseen trajectories, and reconstruct the ground-truth trajectories using the computed latent actions. We benchmark these models against the SOTA DreamDojo IDM baseline on the val/test splits of the in-domains datasets (i.e., datasets seen during training, e.g., Bridge, DROID, and OXE-Mix) and on the following held-out datasets (i.e., unseen data): ``austin\_sailor,'' ``utaustin\_mutex," ``berkeley\_ur5," and ``stanford\_hydra," which include robot morphologies (e.g., the UR5 arm) that were not seen during training.

\begin{figure*}[t]
  \centering
  \includegraphics[width=\linewidth]{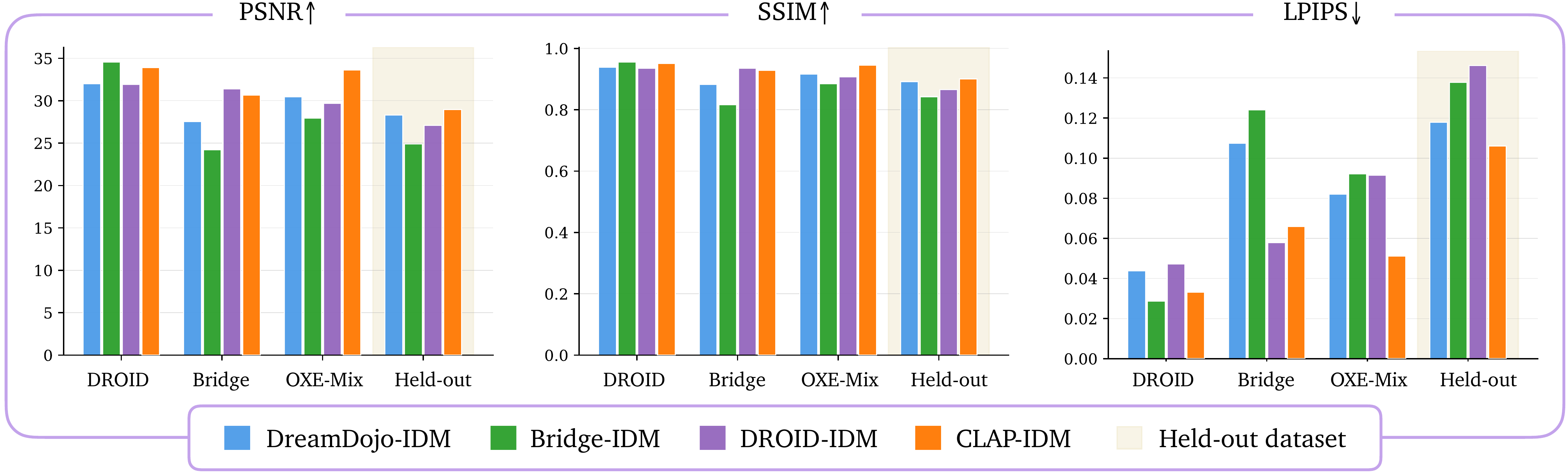}
  \caption{\textbf{Frame-to-frame reconstruction quality.}
  Each LAM is used as an inverse-dynamics model to extract a latent action
  between two consecutive ground-truth frames, which is then decoded to
  reconstruct the next frame. While single-embodiment IDMs perform best on their native distributions, their accuracy degrades sharply out-of-domain. In contrast, \algname's IDM consistently outperforms these off-domain baselines while matching or closely trailing single-embodiment models on their target datasets.}
  \label{fig:lam_f2f}
\end{figure*}

\begin{figure*}[t]
  \centering
  \includegraphics[width=\linewidth]{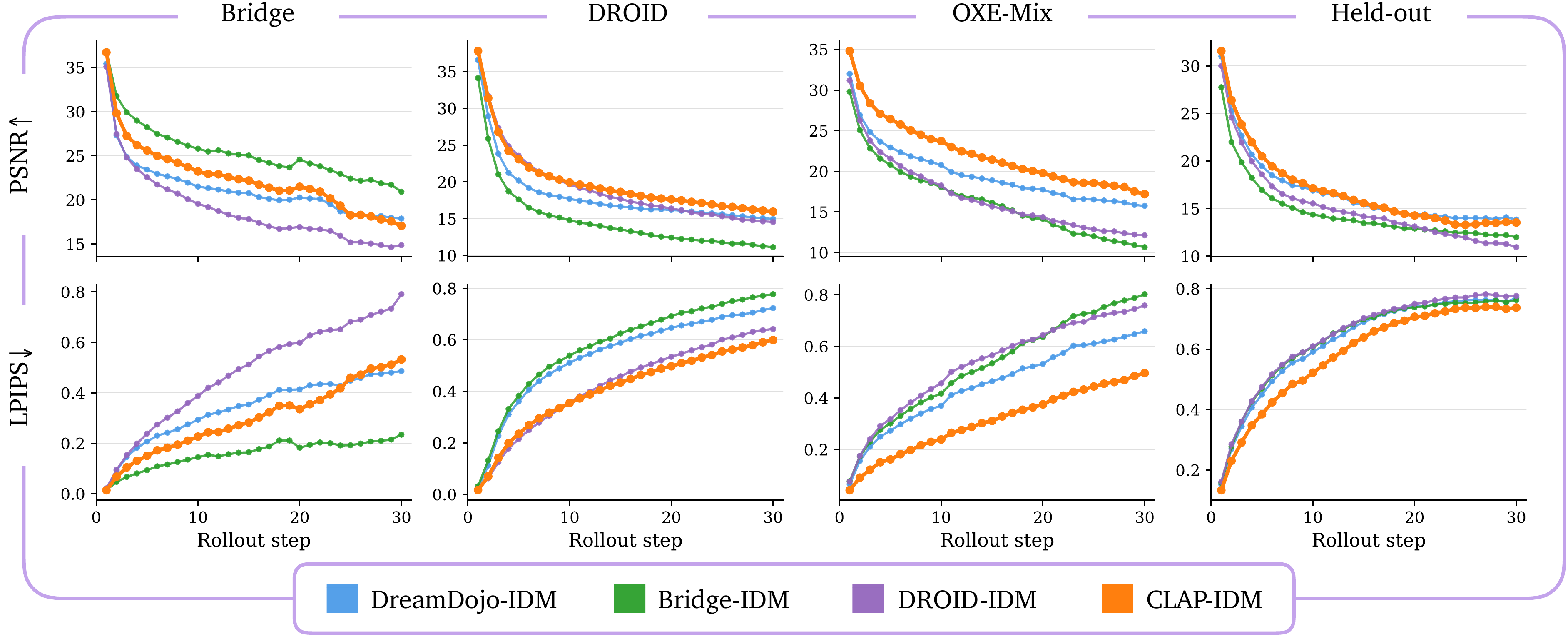}
  \caption{\textbf{LAMs' autoregressive rollouts}.
  Starting from a single ground-truth frame, we rollout each LAM for
  30 steps using its own decoded predictions. Compared to all other methods, \algname's IDM achieves superior generalization beyond the domain of the training datasets.}
  \label{fig:lam_rollout}
\end{figure*}

We evaluate the frame-to-frame prediction accuracy (\Cref{fig:lam_f2f}) and the open-loop rollout accuracy with a skip of two frames (\Cref{fig:lam_rollout}). 
While single-embodiment IDMs perform the best on their target datasets, their performance degrades sharply outside of these datasets, which limits their generalization. In contrast, \algname's IDM generalizes across all datasets and achieves the highest or second-highest perceptual scores, e.g., on the held-out dataset and OXE-mix. These findings hold for both the frame-to-frame reconstruction and open-loop rollout results. In essence, \algname's IDM offers better generalization without compromising in-domain performance, compared to all other methods.

\subsection{Additional Results on Cross-Embodiment Video Modeling}
\label{app:real_world_experiments}

\begin{figure}[t]
    \vspace{-2ex}
    \centering
    \includegraphics[width=0.4\linewidth]{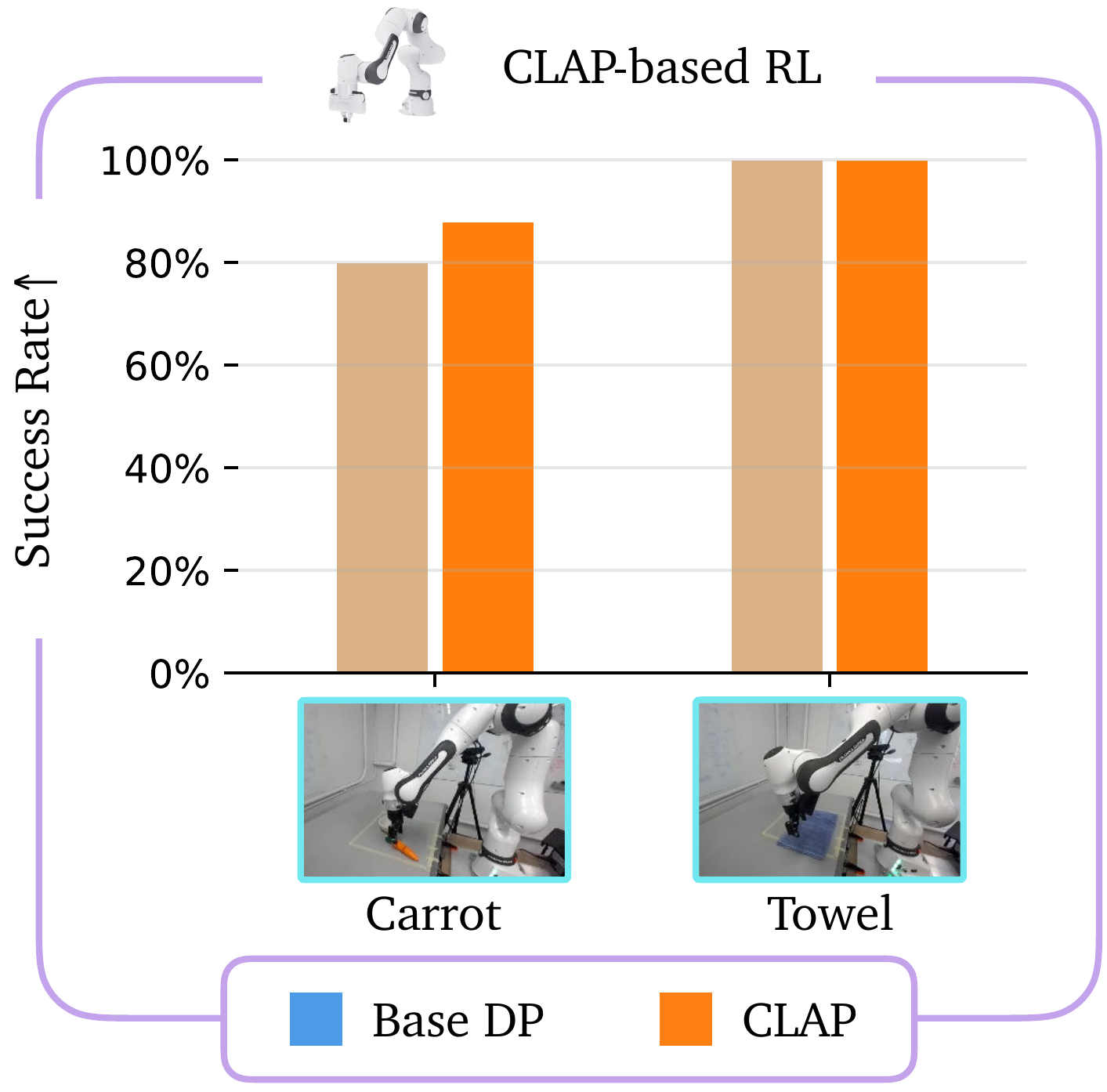}
    \caption{\textbf{Policy Finetuning via RL in \algname.} \algname facilities policy finetuning with reinforcement learning (RL) in video world models. By finetuning base diffusion policies (Base DP), \algname achieves higher success rates in the \emph{carrot} task, while matching the base policy's success rate in the \emph{towel} task.}
    \label{fig:clap_rl}
\end{figure}
\p{Finetuning Robot Policies via Video Model-based Reinforcement Learning}
\algname facilitates policy finetuning in video world models with reinforcement learning (RL). We use the Diffusion Steering via Reinforcement Learning (DSRL) framework~\cite{wagenmaker2025steering} to finetune diffusion policies within \algname's video models and evaluate their success rates in the following real-world tasks: (i) put a carrot in a bowl, and (ii) fold a towel. We predict task progress using a simple convolutional neural network to estimate dense per-frame reward signals in RL. We evaluate the base and finetuned policies across $25$ and $20$ trials in the \emph{carrot} and \emph{towel} tasks, respectively. From~\Cref{fig:clap_rl}, \algname improves the success rate of the base diffusion policy on the \emph{carrot} task via RL in the video model (from $80\%$ to $88\%$), without degrading its performance on the \emph{towel} task. These results demonstrate \algname's potential for cost-effective policy finetuning, circumventing the significant hardware and setup costs associated with alternative pipelines, such as real-world or simulation-based RL. Exploring these directions constitute an exciting avenue for future work.

\p{Comprehensive Experiment Results}
\Cref{tab:cross_vs_single_emb_droid,tab:cross_vs_single_emb_bridge,tab:cross_vs_single_emb_oxe_mix,tab:relative_vs_absolute_action_droid,tab:relative_vs_absolute_action_bridge,tab:relative_vs_absolute_action_oxe_mix,tab:inf_time_single_arm,tab:inf_time_bimanual,tab:few_shot_adapt_pretrained_embodiment_droid,tab:few_shot_adapt_pretrained_embodiment_bridge,tab:few_shot_adapt_novel_embodiment,tab:clap_based_rl} report the results for all experiment.

\begin{table}[t]
    \centering
    \caption{Performance of \algname's cross-embodiment video world models on DROID.}
    \label{tab:cross_vs_single_emb_droid}
    \begin{tabular}{lrrrrr}
    \toprule
    Method & PSNR$\uparrow$ & SSIM$\uparrow$ & LPIPS$\downarrow$ & FVD$\downarrow$ & FID$\downarrow$ \\
    \midrule
    DreamDojo-Human & 12.450 $\pm$ 1.691 & 0.384 $\pm$ 0.080 & 0.681 $\pm$ 0.089 & 116.009 & 241.056 \\
    \alglang & 17.200 $\pm$ 2.740 & 0.692 $\pm$ 0.087 & 0.260 $\pm$ 0.069 & 21.517 & 39.168 \\
    \alglam & 18.859 $\pm$ 2.770 & 0.729 $\pm$ 0.081 & 0.222 $\pm$ 0.068 & 19.059 & 36.863 \\
    \algee & 18.835 $\pm$ 2.837 & 0.734 $\pm$ 0.080 & 0.215 $\pm$ 0.069 & 16.370 & 33.075 \\
    \algcurr & 19.138 $\pm$ 2.671 & 0.744 $\pm$ 0.078 & 0.204 $\pm$ 0.064 & 16.139 & 32.484 \\
    Ctrl-World & 18.928 $\pm$ 2.760 & 0.736 $\pm$ 0.081 & 0.205 $\pm$ 0.067 & 15.591 & 30.543 \\
    \bottomrule
    \end{tabular}
\end{table}

\begin{table}[t]
    \centering
    \caption{Performance of \algname's cross-embodiment video world models on Bridge.}
    \label{tab:cross_vs_single_emb_bridge}
    \begin{tabular}{lrrrrr}
    \toprule
    Method & PSNR$\uparrow$ & SSIM$\uparrow$ & LPIPS$\downarrow$ & FVD$\downarrow$ & FID$\downarrow$ \\
    \midrule
    DreamDojo-Human & 14.376 $\pm$ 1.220 & 0.384 $\pm$ 0.079 & 0.426 $\pm$ 0.089 & 73.442 & 75.060 \\
    \alglang & 20.548 $\pm$ 2.291 & 0.780 $\pm$ 0.049 & 0.142 $\pm$ 0.052 & 17.322 & 25.761 \\
    \alglam & 24.179 $\pm$ 2.032 & 0.827 $\pm$ 0.039 & 0.087 $\pm$ 0.035 & 9.207 & 20.755 \\
    \algee & 23.589 $\pm$ 2.442 & 0.822 $\pm$ 0.043 & 0.091 $\pm$ 0.041 & 10.119 & 19.698 \\
    \algcurr & 23.692 $\pm$ 2.415 & 0.828 $\pm$ 0.041 & 0.088 $\pm$ 0.040 & 9.614 & 19.205 \\
    Bridge-Base & 25.660 $\pm$ 2.294 & 0.858 $\pm$ 0.038 & 0.063 $\pm$ 0.031 & 6.402 & 13.307 \\
    \bottomrule
    \end{tabular}
\end{table}

\begin{table}[t]
    \centering
    \caption{Performance of \algname's cross-embodiment video world models on OXE-Mix.}
    \label{tab:cross_vs_single_emb_oxe_mix}
    \begin{tabular}{lrrrrr}
    \toprule
    Method & PSNR$\uparrow$ & SSIM$\uparrow$ & LPIPS$\downarrow$ & FVD$\downarrow$ & FID$\downarrow$ \\
    \midrule
    DreamDojo-Human & 14.629 $\pm$ 2.793 & 0.421 $\pm$ 0.109 & 0.473 $\pm$ 0.189 & 102.601 & 159.744 \\
    \alglang & 19.705 $\pm$ 2.825 & 0.765 $\pm$ 0.075 & 0.169 $\pm$ 0.077 & 37.010 & 50.749 \\
    \alglam & 22.605 $\pm$ 3.391 & 0.806 $\pm$ 0.071 & 0.125 $\pm$ 0.075 & 20.030 & 43.397 \\
    \algee & 21.665 $\pm$ 2.955 & 0.798 $\pm$ 0.065 & 0.129 $\pm$ 0.068 & 26.199 & 41.854 \\
    \algcurr & 21.898 $\pm$ 2.879 & 0.806 $\pm$ 0.062 & 0.125 $\pm$ 0.064 & 23.202 & 41.218 \\
    \bottomrule
    \end{tabular}
\end{table}

\begin{table}[t]
    \centering
    \caption{Effects of relative and absolute action-space conditioning on cross-embodiment video modeling (DROID).}
    \label{tab:relative_vs_absolute_action_droid}
    \begin{tabular}{lrrrrr}
    \toprule
    Method & PSNR$\uparrow$ & SSIM$\uparrow$ & LPIPS$\downarrow$ & FVD$\downarrow$ \\
    \midrule
    \texttt{\algee-Rel} & 16.758 $\pm$ 2.628 & 0.697 $\pm$ 0.074 & 0.257 $\pm$ 0.066 & 22.113 \\
    \texttt{\algee-Abs} & 18.835 $\pm$ 2.837 & 0.734 $\pm$ 0.080 & 0.215 $\pm$ 0.069 & 16.370 \\
    \texttt{\alglang-Abs} & 16.290 $\pm$ 2.384 & 0.691 $\pm$ 0.075 & 0.276 $\pm$ 0.061 & 26.485 \\
    \texttt{\alglang-Rel} & 17.200 $\pm$ 2.740 & 0.692 $\pm$ 0.087 & 0.260 $\pm$ 0.069 & 21.517 \\
    \bottomrule
    \end{tabular}
\end{table}

\begin{table}[t]
    \centering
    \caption{Effects of relative and absolute action-space conditioning on cross-embodiment video modeling (Bridge).}
    \label{tab:relative_vs_absolute_action_bridge}
    \begin{tabular}{lrrrrr}
    \toprule
    Method & PSNR$\uparrow$ & SSIM$\uparrow$ & LPIPS$\downarrow$ & FVD$\downarrow$ \\
    \midrule
    \texttt{\algee-Rel} & 22.459 $\pm$ 2.399 & 0.841 $\pm$ 0.036 & 0.110 $\pm$ 0.038 & 10.545 \\
    \texttt{\algee-Abs} & 23.589 $\pm$ 2.442 & 0.822 $\pm$ 0.043 & 0.091 $\pm$ 0.041 & 10.119 \\
    \texttt{\alglang-Abs} & 19.317 $\pm$ 2.170 & 0.770 $\pm$ 0.047 & 0.159 $\pm$ 0.050 & 20.356 \\
    \texttt{\alglang-Rel} & 20.548 $\pm$ 2.291 & 0.780 $\pm$ 0.049 & 0.142 $\pm$ 0.052 & 17.322 \\
    \bottomrule
    \end{tabular}
\end{table}

\begin{table}[t]
    \centering
    \caption{Effects of relative and absolute action-space conditioning on cross-embodiment video modeling (OXE-Mix).}
    \label{tab:relative_vs_absolute_action_oxe_mix}
    \begin{tabular}{lrrrrr}
    \toprule
    Method & PSNR$\uparrow$ & SSIM$\uparrow$ & LPIPS$\downarrow$ & FVD$\downarrow$ \\
    \midrule
    \texttt{\algee-Rel} & 20.593 $\pm$ 3.795 & 0.798 $\pm$ 0.084 & 0.157 $\pm$ 0.083 & 25.707 \\
    \texttt{\algee-Abs} & 21.665 $\pm$ 2.955 & 0.798 $\pm$ 0.065 & 0.129 $\pm$ 0.068 & 26.199 \\
    \texttt{\alglang-Abs} & 19.197 $\pm$ 2.783 & 0.765 $\pm$ 0.068 & 0.178 $\pm$ 0.074 & 40.696 \\
    \texttt{\alglang-Rel} & 19.705 $\pm$ 2.825 & 0.765 $\pm$ 0.075 & 0.169 $\pm$ 0.077 & 37.010 \\
    \bottomrule
    \end{tabular}
\end{table}

\begin{table}[t]
    \centering
    \caption{Success rates ($\%$) in inference-time cross-policy planning in single-arm manipulation (DROID).}
    \label{tab:inf_time_single_arm}
    \begin{tabular}{lrrr}
    \toprule
    Task & MolmoAct-2 & $\pi_{0.5}$ & CLAP \\
    \midrule
    Measuring Tape & 75.0 & 40.0 & 80.0 \\
    Fish & 50.0 & 50.0 & 75.0 \\
    Red Lobster & 20.0 & 90.0 & 95.0 \\
    Box & 100.0 & 100.0 & 100.0 \\
    Towel & 100.0 & 100.0 & 100.0 \\
    \bottomrule
    \end{tabular}
\end{table}

\begin{table}[t]
    \centering
    \caption{Success rates ($\%$) in inference-time cross-policy planning in bimanual manipulation (bimanual YAM).}
    \label{tab:inf_time_bimanual}
    \begin{tabular}{lrr}
    \toprule
    Task & MolmoAct-2 & CLAP \\
    \midrule
    Pick Carrot \& Apple & 90.0 & 95.0 \\
    Pick Block \& Avocado & 100.0 & 100.0 \\
    \bottomrule
    \end{tabular}
\end{table}

\begin{table}[t]
    \centering
    \caption{Performance after few-shot adaptation to pretrained embodiments (DROID).}
    \label{tab:few_shot_adapt_pretrained_embodiment_droid}
    \begin{tabular}{lrrrrr}
    \toprule
    Method & PSNR$\uparrow$ & SSIM$\uparrow$ & LPIPS$\downarrow$ & FVD$\downarrow$ & FID$\downarrow$ \\
    \midrule
    \alglang & 19.381 $\pm$ 2.780 & 0.745 $\pm$ 0.080 & 0.198 $\pm$ 0.068 & 15.208 & 30.605 \\
    \alglam & 19.257 $\pm$ 2.552 & 0.742 $\pm$ 0.077 & 0.200 $\pm$ 0.065 & 15.945 & 31.023 \\
    \algee & 19.461 $\pm$ 2.871 & 0.745 $\pm$ 0.082 & 0.191 $\pm$ 0.066 & 14.590 & 27.708 \\
    \algcurr & 19.450 $\pm$ 2.732 & 0.747 $\pm$ 0.078 & 0.194 $\pm$ 0.064 & 14.598 & 30.619 \\
    Ctrl-World & 18.928 $\pm$ 2.760 & 0.736 $\pm$ 0.081 & 0.205 $\pm$ 0.067 & 15.591 & 30.543 \\
    \bottomrule
    \end{tabular}
\end{table}

\begin{table}[t]
    \centering
    \caption{Performance after few-shot adaptation to pretrained embodiments (Bridge).}
    \label{tab:few_shot_adapt_pretrained_embodiment_bridge}
    \begin{tabular}{lrrrrr}
    \toprule
    Method & PSNR$\uparrow$ & SSIM$\uparrow$ & LPIPS$\downarrow$ & FVD$\downarrow$ & FID$\downarrow$ \\
    \midrule
    \alglang & 25.595 $\pm$ 2.165 & 0.860 $\pm$ 0.035 & 0.061 $\pm$ 0.029 & 6.101 & 12.564 \\
    \alglam & 25.821 $\pm$ 2.220 & 0.862 $\pm$ 0.036 & 0.061 $\pm$ 0.029 & 5.779 & 12.822 \\
    \algee & 26.044 $\pm$ 2.189 & 0.865 $\pm$ 0.037 & 0.059 $\pm$ 0.028 & 5.975 & 12.510 \\
    \algcurr & 25.511 $\pm$ 2.332 & 0.858 $\pm$ 0.039 & 0.062 $\pm$ 0.031 & 5.667 & 12.435 \\
    Bridge-Base & 25.660 $\pm$ 2.294 & 0.858 $\pm$ 0.038 & 0.063 $\pm$ 0.031 & 6.402 & 13.307 \\
    \bottomrule
    \end{tabular}
\end{table}

\begin{table}[t]
    \centering
    \caption{Performance after few-shot adaptation to novel embodiments.}
    \label{tab:few_shot_adapt_novel_embodiment}
    \begin{tabular}{lrrrrr}
    \toprule
    Embodiment & PSNR$\uparrow$ & SSIM$\uparrow$ & LPIPS$\downarrow$ & FVD$\downarrow$ & FID$\downarrow$ \\
    \midrule
    Bimanual YAM & 24.310 $\pm$ 3.840 & 0.811 $\pm$ 0.073 & 0.152 $\pm$ 0.072 & 11.519 & 28.564 \\
    G1 humanoid & 15.151 $\pm$ 1.120 & 0.709 $\pm$ 0.053 & 0.215 $\pm$ 0.050 & 16.527 & 19.599 \\
    \bottomrule
    \end{tabular}
\end{table}

\begin{table}[t]
    \centering
    \caption{Success rates ($\%$) after RL-based policy finetuning with \algname.}
    \label{tab:clap_based_rl}
    \begin{tabular}{lrr}
    \toprule
    Task & Base DP & CLAP \\
    \midrule
    Put Carrot in Bowl & 80.0 & 88.0 \\
    Fold Towel & 100.0 & 100.0 \\
    \bottomrule
    \end{tabular}
\end{table}

\clearpage

\section{Nuanced Summary}
\label{sec:app_nuanced_summary}

This section discusses the nuances surrounding \algname's novelty, broader potential impact, and scope of claims.

\listofquestions
\vspace{1.5em}
\hrule
\vspace{1.5em}

\section*{Detailed Q\&A}

\begin{description}[leftmargin=3.5em, style=nextline]
    \qitem[qa:novelty]{\algname uses an old video backbone (SVD, 2023), a pretrained language encoder (CLIP), simple MLP action encoders, and does not introduce a new dataset. Is it novel?}
    {Not quite. By that same token, current state-of-the-art action-conditioned video models are largely derivative. \algname's novelty lies in introducing a new paradigm for action-conditioned video modeling rooted specifically in cross-embodiment learning.}
    
    \qitem[qa:why-not-newer-backbones]{Why does \algname use SVD and not newer backbones like Wan 2.2, and does it generalize to these backbones?}
    {\algname adopts an SVD backbone to enable thorough experimentation across action, image, and history conditioning schemes and data mixtures at a minimal cost. While newer backbones (such as Wan 2.2) yield higher-quality video, they require roughly 10 days of training per run on an H200 node compared to SVD's 2 days, representing prohibitive costs for all but industry-tier budgets. Moreover, SVD enables inference on budget-friendly consumer-grade hardware unlike many newer backbones. Crucially, \algname generalizes seamlessly to these newer backbones because it is not tied to any architecture-specific features.}
    
    \qitem[qa:diffusion-vs-flow-matching]{What motivated the choice of diffusion over alternative frameworks like flow-matching?}
    {\algname builds directly on SVD's native diffusion framework for stable integration and rapid iteration. However, future work will adopt flow matching for accelerated inference.}

    \qitem[qa:consumer-gpu]{Can \algname run on consumer-grade GPUs?}
    {Yes, \algname requires under $12$~GB of VRAM at inference, which easily fits on consumer hardware like the RTX 3060.}

    \qitem[qa:realtime-capability]{Does \algname run in real-time (at least $10$~Hz)?}
    {No, \algname falls short of $10$~Hz. Its inference latency is hardware-dependent, averaging $3.24\text{s}$ on the A100-PCIe-40GB, $2.88\text{s}$ on the A100-SXM4-80GB, and $1.49\text{s}$ on the H200 for a standard configuration with $11$ frames and $25$ denoising steps, averaged over $20$ trials.}
    
    \qitem[eq:internet-scale-data]{Is the training scale of \algname comparable to GPT, Claude, or Gemini, and can it realistically handle internet-scale video data?}
    {No, \algname is trained on significantly smaller datasets than foundation models like GPT, Claude, or Gemini. However, \algname's architecture natively supports internet-scale data, though training at that scale remains resource-intensive. We welcome collaborations to scale \algname to internet-scale datasets. In preliminary experiments, we observe that \algname scales favorably with both model size and data volume.}
    
    \qitem[qa:upper-limit-of-cross-embodiment-learning]{Do the benefits of cross-embodiment learning plateau?}
    {Ultimately, but we see no evidence to suggest that this plateau is near.}
    
    \qitem[qa:comparison-simulators]{Does \algname outperform physics-based simulators, and is its output high-fidelity?}
    {Like other video models, \algname is not yet a replacement for well-tuned physics-based simulators, but it offers strong potential to eventually supersede them. Moreover, \algname's predictions achieve a level of fidelity that approaches or surpasses existing state-of-the-art single-embodiment video models in challenging environments.}
    
    \qitem[qa:better-than-single-embodiment]{Does \algname consistently outperform all single-embodiment models?}
    {No, \algname does not universally outperform single-embodiment models. Like other cross-embodiment or foundation models (e.g., GPT), it can occasionally underperform domain-specific models tailored to a single embodiment.}

    \qitem[qa:compatibility-mobile-robots]{Will \algname work effectively with all robots, e.g., mobile robots?}
    {Not universally. Performance depends on \algname's scale and distribution coverage, but accuracy improves as target robot platforms grow closer to the training distribution. Moreover, further adaptation to floating reference frames may be required.}
    
    \qitem[qa:hallucination]{Does \algname hallucinate?}
    {Yes, like existing SOTA video models, \algname is prone to hallucinations at the boundaries of its training distribution, presenting an exciting direction for future work on uncertainty quantification and hallucination mitigation.}
    
    \qitem[qa:better-action-representation]{Which action representation yields optimal performance?}
    {No single action representation universally outperforms the others; performance depends on a nuanced trade-off where language actions minimize the domain gap for foundation models but lack precision, end-effector poses offer high precision but require labeled video data, and latent actions unlock unlabeled video data usage while introducing a deployment domain gap.}
    
    \qitem[qa:meaningful-latent-actions]{What is the actual utility of learned latent actions?}
    {Learned latent actions enable cross-embodiment learning on unlabeled internet-scale videos by bypassing the need for manual action annotations.}
    
    \qitem[qa:comparisons-among-action-rep]{Is it a fair comparison between language, end-effector poses, and latent actions?}
    {Whether a direct comparison is fair depends on the goal, as \algname focuses on cross-embodiment modeling rather than benchmarking representations. Forcing a data-matched comparison would require training all action representations on the same data, which would block the use of unlabeled (internet-scale) data and defeat the framework's core objective.
    }
    
    \qitem[qa:incapable-language-actions]{Are language actions fundamentally poor at video prediction?}
    {Language actions are not inherently poor at video prediction; rather, existing language encoders are simply not optimized for processing numerical data, creating an artificial gap that limits their overall expressiveness.}

    \qitem[qa:single-emb-vs-post-trained]{Is it fair to compare single-embodiment models to post-trained models?}
    {Yes, this comparison is fair given \algname's goal of providing a straightforward way to adapt cross-embodiment models to a target embodiment. This ensures that both the single-embodiment and post-trained models operate under the same effective user-side training budget.}
    
    \qitem[qa:better-than-robot-policies]{Does \algname always beat $\pi_{0.5}$ and MolmoAct-2 via inference-time planning?}
    {Not always. While inference-time planning allows \algname to explore different action proposals from various policies to boost performance, it can still fail due to issues like video or reward prediction hallucinations.}
    
    \qitem[qa:perceptual-vs-object-metrics]{Why not use object-centric metrics like object masks and poses extracted with an inverse dynamics model (IDM) or a pose tracker instead of perceptual metrics?}
    {While object-centric metrics can offer fine-grained assessments, state-of-the-art IDMs and pose trackers are prone to hallucinating outputs across diverse tasks, which can compromise evaluation integrity. To avoid this, \algname uses standard perceptual metrics common in state-of-the-art video models, leaving improved evaluation metrics as an open area of research.}
    
    \qitem[qa:baseline-comparisons]{Why not compare against more video model baselines (backbones)?}
    {First, no off-the-shelf cross-embodiment action-conditioned video models exist. To eliminate confounding variables such as model size or pretraining data, all models are trained on the same backbone derived from the state-of-the-art DROID model. Furthermore, in the Bridge environment, this baseline already outperforms existing state-of-the-art methods.}
    
    \qitem[qa:real-world-video-model-comparisons]{Why not compare against other video models in the planning and RL experiments?}
    {The primary goal of the planning and RL experiments is to demonstrate real-world zero-shot generalization, making additional video model comparisons unnecessary for a few reasons. First, applying video models to planning or RL is already well-established, predating current state-of-the-art video models. More importantly, performance differences between state-of-the-art video models rarely manifest in real-world deployment, typically appearing only in contrived edge cases that current robot policies are not even capable of operating within.}
    
    \qitem[qa:comparison-for-novel-embodiments]{Why not compare against other baselines when adapting to novel embodiments?}
    {These experiments focus on demonstrating seamless adaptation to novel embodiments. While fine-tuning already proves that \algname provides strong priors for in-domain settings, cross-domain training (e.g., from a robot-data-free SVD backbone to robot video prediction) incurs no performance penalty compared to training from scratch. In fact, within a single gradient step, our adapted models generate meaningful video predictions, unlike models trained from raw SVD backbones. Additionally, no open-source video model baselines currently exist for bimanual YAM and the G1 Humanoid.}
    
    \qitem[qa:action-reconstruction-idm-metric]{Why not use action reconstruction as the metric when comparing the IDMs instead of future-frame reconstruction?}
    {The main goal of the IDM is to extract actions from unlabeled videos. Because ground-truth actions are unavailable in these datasets, \algname relies on future-frame reconstruction to assess the accuracy of the latent actions.}
    
    \qitem[qa:reward-model-rl-experiments]{Why did you use a simple GPT-based reward model in the RL experiments?}
    {As noted earlier, ablating the reward model is not central to the goal of the RL experiments. In practice, any feasible reward model --- even a human operator --- can be swapped into the deployment pipeline. The GPT-based reward model was simply sufficient for our setup.}
    
    \qitem[qa:bimanual-stress-test]{Is evaluating single-arm \algname models on bimanual robots informative?}
    {It is informative as a stress test, pushing the single-arm \algname model outside its standard operational domain by introducing two robots into a single scene. By managing the action-conditioning gap (accounting for the difference between the 7-dimensional model and the 14-dimensional bimanual setup), we can effectively deploy and evaluate the model. Naturally, for optimal performance, the model should be adapted to the target embodiment, as demonstrated in our experiments.}

\end{description}

}

\end{document}